\documentclass[11pt]{article}

\usepackage[margin=1in]{geometry}
\usepackage{graphicx}
\usepackage{booktabs}
\usepackage{amsmath,amssymb}
\usepackage{hyperref}
\usepackage{natbib}
\usepackage{authblk}
\usepackage{microtype}
\usepackage[T1]{fontenc}
\usepackage[utf8]{inputenc}
\usepackage{color}

\usepackage{enumitem}
\usepackage{booktabs}
\usepackage{setspace}
\usepackage{xcolor}

\title{The Human Flourishing Geographic Index: A County-Level Dataset for the United States, 2013--2023}

\author[1]{Stefano M.\ Iacus}
\author[2]{Devika Jain}
\author[1,5]{Andrea Nasuto}
\author[3]{Giuseppe Porro}
\author[4]{Marcello Carammia}
\author[6]{Andrea Vezzulli}

\affil[1]{Institute for Quantitative Social Science, Harvard University, Cambridge, MA 02138, USA}
\affil[2]{Center for Geographic Analysis, Harvard University, Cambridge, MA 02138, USA}
\affil[3]{Department of Law, Economics and Culture, University of Insubria, Como, Italy}
\affil[4]{Department of Political and Social Sciences, University of Catania, Italy}
\affil[5]{Geographic Data Science Lab, University of Liverpool, UK}
\affil[6]{InsIDE Lab - Department of Economics, University of Insubria, Varese, Italy}

\date{\today}

\begin{document}
\maketitle

\begin{abstract}
Quantifying human flourishing—a multidimensional construct including happiness, health, purpose, virtue, relationships, and financial stability—is critical for understanding societal well-being beyond economic indicators. Existing measures often lack fine spatial and temporal resolution. Here we introduce the Human Flourishing Geographic Index (HFGI), derived from analyzing approximately 2.6 billion geolocated U.S. tweets (2013–2023) using fine-tuned large language models to classify expressions across 48 indicators aligned with Harvard’s Global Flourishing Study framework plus attitudes towards migration and perception of corruption. The dataset offers monthly and yearly county- and state-level indicators of flourishing-related discourse, validated to confirm that the measures accurately represent the underlying constructs and show expected correlations with established indicators. This resource enables multidisciplinary analyses of well-being, inequality, and social change at unprecedented resolution, offering insights into the dynamics of human flourishing as reflected in social media discourse across the United States over the past decade.
\end{abstract}


\section*{Corresponding author(s)}
Stefano M.\ Iacus (\texttt{siacus@iq.harvard.edu}); Devika Jain (\texttt{kakkar@fas.harvard.edu}).

\clearpage


\clearpage

\section{Background \& Summary}
Quantifying human flourishing is essential to understanding societal progress beyond economic growth or material wealth. Traditional indicators such as GDP or income capture production and consumption but overlook the multidimensional nature of human well-being---including purpose, relationships, autonomy, health, and meaning. Existing well-being reports, such as the World Happiness Report \cite{Helliwell2025}, provide valuable global or national benchmarks but typically lack fine-grained spatial and temporal resolution. Yet flourishing is inherently contextual and dynamic, shaped by local social, economic, and environmental conditions that evolve over time. Measuring flourishing at multiple geographic levels, such as states and counties, enables researchers to examine the spatiotemporal dynamics of well-being, uncover regional disparities, and identify the local and temporal factors that foster or hinder human thriving. Spatially and temporally explicit indicators also allow integration with other public data sources, including socioeconomic statistics, health and demographic surveys, and CDC (Centers for Disease Control and Prevention)  datasets, thereby supporting cross-domain analyses and evidence-based policy interventions.

Evaluating individual and collective human flourishing has emerged as a prominent and ambitious goal among social scientists and policymakers in recent years (for a survey see \citep{rule2024flourishing}). Among these attempts, Harvard’s Human Flourishing Program \citep{vanderweele2017promotion} is dedicated to studying and promoting human flourishing across diverse domains of life. When thinking about flourishing in a pluralistic or multicultural context it is helpful to consider five domains of human life: \textit{i)} happiness and life satisfaction, \textit{ii)} mental and physical health, \textit{iii)} meaning and purpose, \textit{iv)} character and virtue, \textit{v)} close social relationships; and \textit{vi)} a sixth ``means'' domain of  material and financial stability.

The Global Flourishing Study (GFS) \citep{gfs} is a five-year longitudinal study originated within the framework of the Human Flourishing Program, involving approximately 200{,}000 participants from over 20 countries and territories. Work, education, family, and religious community also give rise to a set of opportunities and challenges for flourishing. These social connections aspects were natural targets for our well-being indicators indicators based on social media data.

The Human Flourishing Geographic Index (HFGI) presented here, inspired by Harvard’s Human Flourishing Program but not part of it, aims to capture expressions of human flourishing through the analysis of 2.6 billion geolocated tweets in the US from the Harvard CGA’s Geotweet Archive v2.0 \citep{geotweet2016}. The analysis of tweets is performed by applying specifically fine-tuned open-source large language models (LLMs) at scale \cite{finetuning2024}. The source dataset spans January 2013 to June 2023.

We extracted a few questions from the Human Flourishing and GFS framework to generate 46 human flourishing indicators, plus two additional measures: \textit{i)} attitudes towards migration \cite{learningtopic} and \textit{ii)} perception of corruption, for a total of 48 indicators.  Though the raw data have intra-daily frequency, our indicators are calculated at both monthly and yearly frequency, and at both county- and state-level, to protect privacy and ensure sufficient volumes for robust aggregation. Table~\ref{tab:flourishing_GFS} shows a mapping between our indicators and the questions in the GFS survey while Table~\ref{tab:flourishing_dimensions} completes the description of all indicators.

Traditionally, surveys have been the primary method for eliciting individuals’ reflections on their lives, emotions, and attitudes \cite{gallup2018healthindex,ess2025,wvs2025}. However, declining response rates  \cite{nayak2019strengths}, observer effects \citep{deaton2016context}, and high costs limit high-frequency measurement. In this context, the Internet \cite{greyling2025happiness} and social networking platforms \citep{iacusporro2021book,iacusporro2022significance} provide a complementary lens: individuals spontaneously share reactions to events without prompting from researchers, enabling measurement at scale with fewer demand effects. Moreover, as also summarized by OECD in their guidelines \citep{oecd2013guidelines}, the multidimensionality of subjective well-being (hedonic, evaluative, and eudaimonic dimensions) has to be central to our extraction strategy. Following prior critiques of social-media SWB indices, we consistently interpret HFGI as an expression propensity rather than population prevalence and evaluate its behavior against external series (sentiment, CDC mental-health indicators, CPI).

\section{Methods}

\subsection{Twitter data collection}\label{sec:datacollection}
The Human Flourishing Geographic Index is derived from the Harvard CGA Geotweet Archive v2.0 \cite{geotweet2016}, a large-scale repository of geo-tagged tweets collected globally from 2010 to July 12, 2023. The archive contains around 10 billion tweets across multiple languages and countries. Tweets are collected via the Twitter Streaming API \citep{xstream} with only tweets carrying explicit spatial metadata retained (about 1--2\% of all tweets on average). The entire collection spans January 2010 to July 12, 2023, when Twitter discontinued Academic API access. U.S.-specific tweet counts are reported in Supplementary Table~\ref{tab:tweet_volume_us_share}.

The archive includes 24 fields capturing spatial, temporal, and non-spatial attributes (Supplementary Table~\ref{tab:dbase}). Two forms of spatial information are associated with tweets: precise GPS coordinates (when available), or a named ``Place'' with a bounding box; when only a Place is available, we use the centroid of its bounding box and store a spatial error derived from the bounding box area. Based on these attributes, we derive three variables: (1) latitude/longitude, (2) a GPS flag, and (3) a spatial error estimate.

\begin{figure}[ht]
\centering
\includegraphics[width=0.88\linewidth]{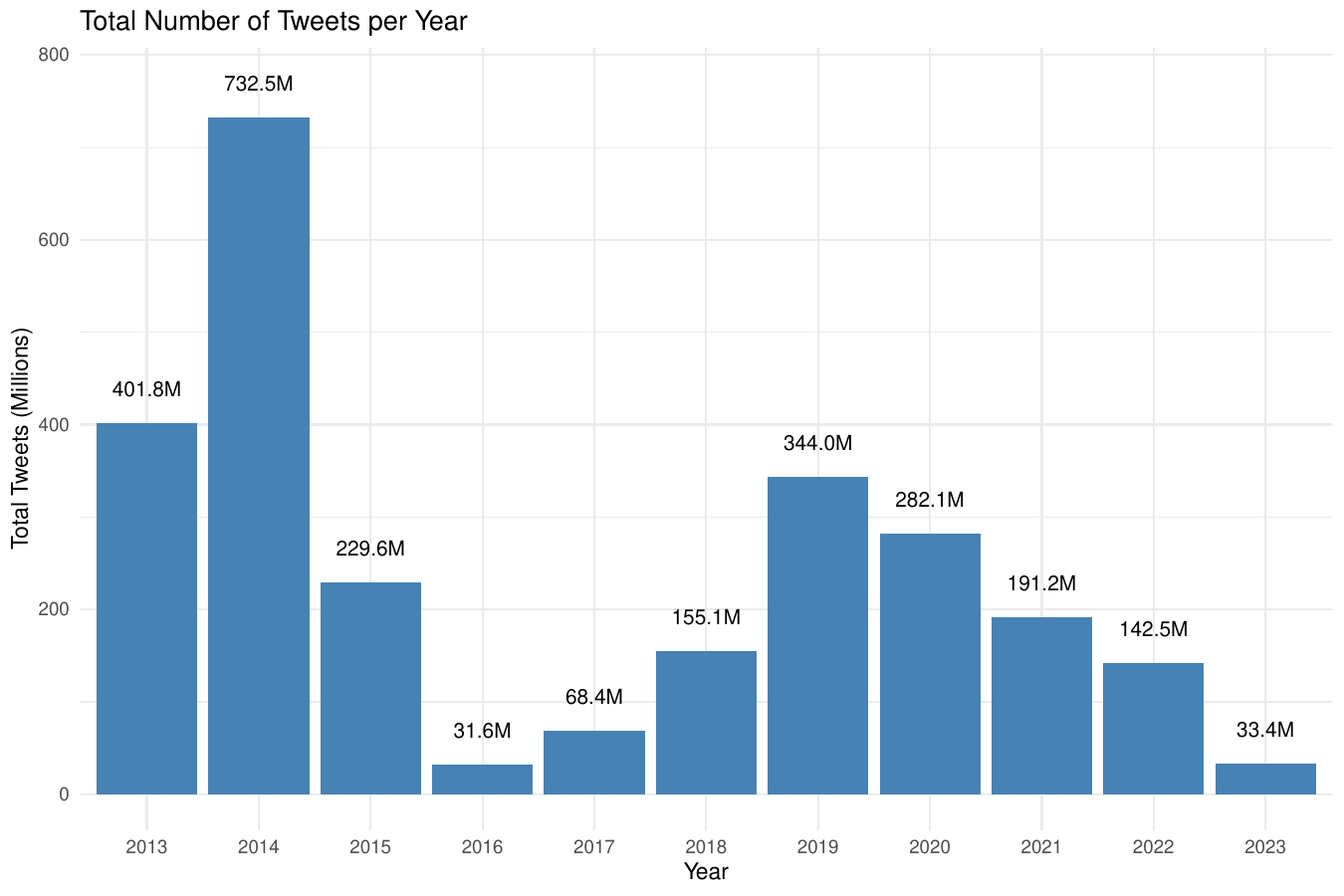}
\caption{Tweet volume over time in the 2.6B dataset for United States.}
\label{fig:actual_tweets}
\end{figure}

For this study, we focused on 2.6 billion tweets originating from the United States, geolocated and enriched with census tract variables using the U.S. Census Bureau’s TIGER/Line Shapefiles (2020) \citep{tiger2020}. The distribution over time of the actual volume of tweets processed in our analysis is presented in Figure~\ref{fig:actual_tweets}.

\subsection{Extracting the flourishing dimensions}
Our textual analysis goes beyond classical sentiment analysis and simple polarity detection. We selected 46 questions from the GFS questionnaire and reformulated them into concise prompts for a large language model (LLM). 
In our analysis, a LLM is asked about the presence of each of the 46 dimensions in a tweet and, if present, to rate the tweet ``low" or ``high''. For example, as for the happiness dimension: happiness = ``high" means true happiness, but if happiness = ``low" it corresponds to sadness. During the analysis, we realized that the dimension is  present either with mild or strong intensity. Therefore, we introduced a three-level scale (``low", ``medium", ``high'') and, after classification, we recoded each dimension to these numeric values: $-1$ (\textit{negative}), $+0.5$ (\textit{somewhat positive}), $+1$ (\textit{positive}) and $0$ (\textit{not-present}).
The example codebook is given as part of the supplementary material.

\subsection{Large Language Model classification}

For this task, we fine-tuned the Llama~3.2~3B model \cite{llama32_3b}. The choice of a relatively small model reflects a deliberate trade-off between classification accuracy and the scalability required to process 2.6~billion tweets. It is well established \cite{finetuning2024} that, with appropriate fine-tuning, smaller models can perform competitively with their larger counterparts. Indeed, models within the same family (e.g., the Llama series) are typically trained on similar or even identical corpora. The number of parameters primarily determines the model’s ability to generalize across tasks and respond to a broader range of inputs. Fine-tuning, in contrast, does not introduce new knowledge but rather helps the model specialize and focus on a specific task. 
Another reason why we chose version 3.2 is that this model officially supports 8 languages (English, German, French, Italian, Portuguese, Hindi, Spanish, and Thai) and this is key for our application which aims at scaling further this work to the rest of the global geo Twitter archive. We also validate \cite{learningtopic} how fine-tuning works across languages in a separate work and we found that fine-tuning helps the model to scale to  languages beyond the official 8 ones with little or none examples. In practice, fine-tuning helps the model to understand the task more than the language as, in fact, the model already know how to map languages through their embedding space.

\subsubsection{Language stability and semantic robustness.}
Unlike term-based NLP sentiment extractors that require periodic vocabulary updates to address lexical drift, our approach relies on a large language model fine-tuned for conceptual understanding of well-being dimensions. Because Llama~3.2~3B has been pretrained on a vast multilingual corpus encompassing (15 trillion tokens) almost exhausting  the entire web content and more, and fine-tuning adjusts task focus rather than vocabulary, it generalizes across new slang, orthographic variants, and evolving idioms without retraining. This semantic adaptability removes the need for ``word-shift'' calibration used in traditional NLP pipelines and ensures consistency of interpretation over the 2013–2023 period and beyond.

\subsection{Building the indicators}

After each tweet is passed to the LLM for classification, it is labeled with ``low'', ``medium'' or ``high'' for each dimension that is identified as present in the text. Then, for each dimension and tweet, the textual labels are transformed into $-1$ for the ``low'' category, $+0.5$ for the category ``medium'', $+1$ for ``high'', and finally as $0$ if the dimension is absent. In the end, all tweets are coded with one of these values $(-1, 0, 0.5, +1)$. 

We assign zeros to all dimensions not detected in a tweet in order to maintain a fixed-length representation across the 46 (or 48) flourishing dimensions. This design choice allows us to exploit highly optimized SQL-style aggregation operations in \texttt{DuckDB} \cite{duckdb2019}, such as vectorized summations and group-by computations, which are critical for processing data at the scale of 2.6~billion tweets. However, during the computation of indicators, zeros are excluded from the denominators so that the resulting averages and standard deviations are calculated only over the (valid number of) tweets in which the given dimension is actually present.

Let's denote by $D_{i,k}$ the value in which tweet $i=1, \ldots, 2.6\textrm{B}$ is coded along dimension $k = 1,\ldots,48$. At this point, for each dimension $k$, the codes are aggregated by summing first by census-level area $A$ for a given day $d$:
$$
C_{k,d}^A = \sum_{i\in {\cal T}^A_d} D_{i,k}
$$
where ${\cal T}^A_d$ is the set of indexes of tweets posted on day $d$ and geo-localized to census area $A$. 

We then aggregate these statistics further at the county and state levels, either at monthly or yearly frequency, i.e.,
$$
C_{k,m}^c = \sum_{d\in m} \sum_{A\in c} C_{k,d}^A
\quad
\text{ and }
\quad 
C_{k,m}^S = \sum_{d\in m} \sum_{c\in S} C_{k,m}^c,
$$
where $c$ denotes a county, $S$ a U.S. state, and $m$ the set of days within a given month. Similarly, for the yearly counts for year $y$:
$$
C_{k,y}^c = \sum_{m\in y} C_{k,m}^c
\quad
\text{ and }
\quad 
C_{k,y}^S = \sum_{m\in y} C_{k,m}^S.
$$

To obtain the indicators, we divide each sum by the valid number of tweets $n_{k,t,g}$ for a particular time $t$ and geographical resolution $g$ for a given dimension $k$, that is,
$$
I_{k,m}^c = \frac{C_{k,m}^c}{n_{k,m,c}}, \quad
I_{k,m}^S = \frac{C_{k,m}^S}{n_{k,m,S}}, \quad
I_{k,y}^c = \frac{C_{k,y}^c}{n_{k,y,c}}, \quad
I_{k,y}^S = \frac{C_{k,y}^S}{n_{k,y,S}}.
$$

All the above indicators $I_{k,t}^g$ vary in $[-1,+1]$, representing the average polarity or intensity of each flourishing dimension at the chosen temporal and spatial resolution. Along with the indicators, we also calculate the standard deviations of these averages, at the corresponding temporal and geographical resolution, denoted by $\sigma_{k,t}^g$, for either $g=c$ (county) or $g=S$ (state), and $t=m$ (monthly) or $t=y$ (yearly).

We further compute a salience measure (see Section~\ref{sec:salience}), corresponding to the share of tweets classified as expressing a given dimension. While this does not capture cognitive or linguistic salience in the strict sense \cite{giora2003salience,schmid2016salience}, it provides an estimate of the relative prevalence of each dimension in the dataset.

\subsection{Computing and Scalability Aspects}
To classify 2.6 billion tweets we take advantage of several high performance computer clusters. We used mainly NVIDIA A100/H100 gpus, running quantized fine-tuned Llama models for a total amount of 1,045,048 hours. We mostly use spare GPU cycles, meaning that we used GPUs only when the owners of those GPUs were not using them for their own research. For this reason we had to apply several check-pointing strategies and parallel distribution tricks as, in this context, GPUs could have been requested by the owners at any time, resulting in the killing of our processes.
We also made use of scalable database engines, like \texttt{DuckDB} \cite{duckdb2019}, to store temporary calculations and execute vectorized computations on disk rather than in memory. Indeed, such systems  operate on files rather than on a centralized databases structure that do not perform well in a HPC cluster environment.
We made our scripts and models portable so that we were able to use different clusters (FAS-RC, Kempner Institute and NSF-ACCESS facilities).
Both FAS-RC and Kempner Institute GPUs are hosted at the Massachusetts Green High Performance Computing Center (MGHPCC), which operates on 100\% carbon-free electricity and employs high-efficiency cooling systems that minimize water consumption. This infrastructure significantly reduces the environmental footprint of our large-scale computations compared to commercial AI providers \cite{learningtopic}.

\section{Data Records}
The final dataset of multi-level indices (county, state) and multi-temporal resolution (monthly, yearly) is available at Harvard Dataverse \cite{HFGI} through this direct link \url{https://doi.org/10.7910/DVN/T39JBY}. Table~\ref{tab:county_year_schema} shows the schema of the County-Year table representing the indicators $I_{k,y}^c$ and their standard deviations $\sigma_{k,y}^c$ while
Table~\ref{tab:county_month_schema} contains the County–Month schema, 
Table~\ref{tab:state_year_schema}  the State–Year schema, and
Table~\ref{tab:state_month_schema}  the State-Month schema. 

As for the notation, the term \texttt{stat\_se} was used internally to map a generic \textit{statistical error}. In the end, we found more appropriate to use the standard deviation, therefore \texttt{stat\_se} is the actual standard deviation of each indicator and not the standard error as the clumn name may actually suggest. For compatibility with prior internal releases of the tables, we keep our naming convention. Users of this dataset may obtain a standard error by dividing this value by the square root of \texttt{validtweets}.

\begin{table}[htbp]
\centering
\caption{Schema for \texttt{flourishingCountyYear.(csv|parquet)} — County-level, yearly records (one row per \{county, year, variable\}).}
\label{tab:county_year_schema}
\begin{tabular}{p{3.3cm} p{2.6cm} p{9cm}}
\hline
\textbf{Field} & \textbf{Type} & \textbf{Description} \\
\hline
\texttt{variable} & \texttt{STRING} & Dimension name (e.g., \texttt{happiness}, \texttt{optimism}, \texttt{loneliness}, \texttt{jobsat}, etc.). \\
\texttt{stat} & \texttt{FLOAT} & Indicator value in $[-1,+1]$ (conditional mean over tweets where the dimension is present). May be empty (NA). \\
\texttt{stat\_se} & \texttt{FLOAT} & Standard deviation of \texttt{stat} (if computed). May be empty (NA). \\
\texttt{salience} & \texttt{FLOAT} & Share/proportion of applicable tweets among all tweets for the cell (0–1). \\
\texttt{ntweets} & \texttt{INT} & Total tweets for the cell (all tweets, regardless of variable presence). \\
\texttt{validtweets} & \texttt{INT} & Tweets where the given \texttt{variable} is present (denominator for conditional mean). \\
\texttt{natweets} & \texttt{INT} & Tweets where the \texttt{variable} is absent (often \texttt{ntweets - validtweets}). \\
\texttt{FIPS} & \texttt{STRING} & 2-digit state FIPS code (zero-padded). Example: \texttt{"06"}, \texttt{"36"}. \\
\texttt{county} & \texttt{STRING} & 3-digit county code within state (zero-padded). Example: \texttt{"083"}, \texttt{"119"}. \\
\texttt{StateCounty} & \texttt{STRING} & 5-digit county FIPS (= \texttt{FIPS} $\Vert$ \texttt{county}); example: \texttt{"06083"}, \texttt{"36119"}. \\
\texttt{year} & \texttt{INT} & Calendar year of aggregation (e.g., \texttt{2010}). \\
\hline
\end{tabular}
\end{table}


\clearpage 

\section{The Spatial Distribution of the HFGI Data}
As discussed earlier (see also Figure~\ref{fig:actual_tweets}), the temporal distribution of tweets is irregular, yet the spatial distribution of the indicators reveals even more informative patterns.
Because social media are typically more \emph{reactive} to events than \emph{agenda-setting} in shaping civic discourse \citep{mccombs1972agenda,mccombs2005agenda,parmelee2012agenda,ceron2016agenda},
relative variation across time is an expected phenomenon (beyond the technical factors inherent to the platform discussed in Section~\ref{sec:datacollection}.)
In this section, we highlight the spatial distribution of the HFGI dimensions, emphasizing that their persistent geographic structure is unlikely to arise from short-term event responsiveness alone.

\subsection{Salience of Flourishing Dimensions}
\label{sec:salience}

In the Human Flourishing Geographic Index (HFGI), we define \textit{salience} as the relative prevalence of a flourishing dimension in geolocated social media discourse.  
Operationally, salience is computed for each \{geography, time, dimension\} cell as the fraction of tweets in which that dimension is expressed:
\[
\text{salience}_{g,t,d} = \frac{\texttt{validtweets}_{g,t,d}}{\texttt{ntweets}_{g,t}},
\]
where \texttt{validtweets} denotes the number of tweets containing expressions related to the given dimension, and \texttt{ntweets} is the total number of geolocated tweets for that cell.  
This measure therefore captures the \emph{relative frequency of topical expression} rather than cognitive or attentional salience in the strict psychological sense \cite{giora2003salience,Schmid2016_Salience}.  
It reflects how often people publicly engage with a given aspect of flourishing within the broader social discourse, conditioning on the universe of messages produced in the same spatial and temporal context.

\begin{figure}[ht!]
\centering
\includegraphics[width=0.9\textwidth]{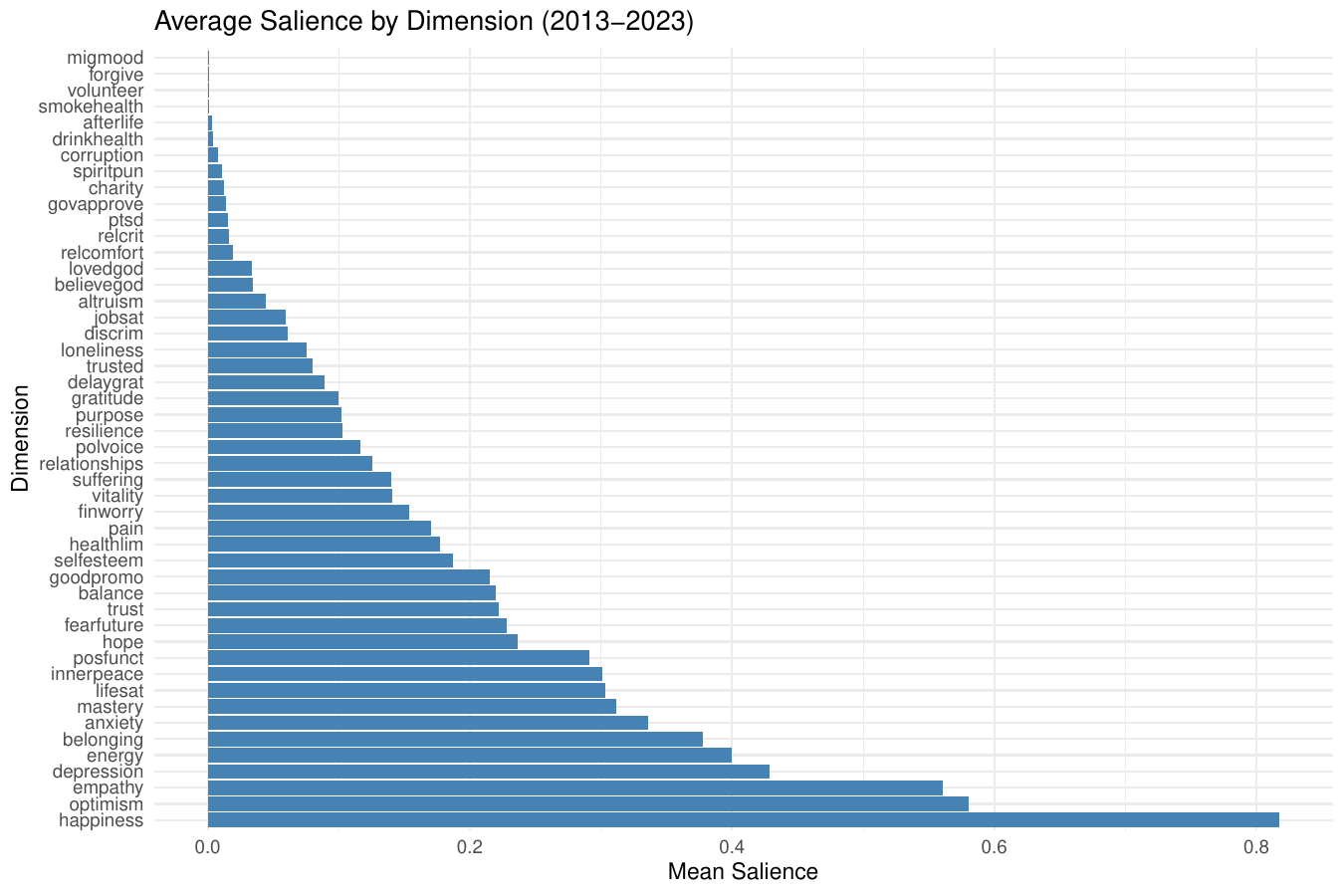}
\label{fig:salience}
\end{figure}

Figure~\ref{fig:salience} shows the average salience of each dimension between 2013 and 2023.  
Indicators related to subjective well-being and emotion (\textit{happiness}, \textit{optimism}, \textit{empathy}, \textit{depression}) exhibit the highest salience values, indicating that these topics are most frequently discussed in everyday online conversation.  
By contrast, dimensions linked to civic, moral, or transcendent themes (\textit{forgiveness}, \textit{volunteering}, \textit{charity}, \textit{afterlife}) appear much less frequently, reflecting their narrower discursive footprint on social media.  
This pattern is consistent with prior findings that affective and experiential concepts dominate user-generated content, while abstract or moral constructs are more sporadic \cite{MarwickBoyd2011_ContextCollapse,Krumpal2013_SDB}.

The distribution of salience across dimensions thus provides an empirical lens on which aspects of human flourishing are more publicly articulated and which remain relatively latent in online expression.

\subsection{Geographic patterns in salience}

\begin{figure}[ht!]
\centering
\includegraphics[width=0.4\textwidth]{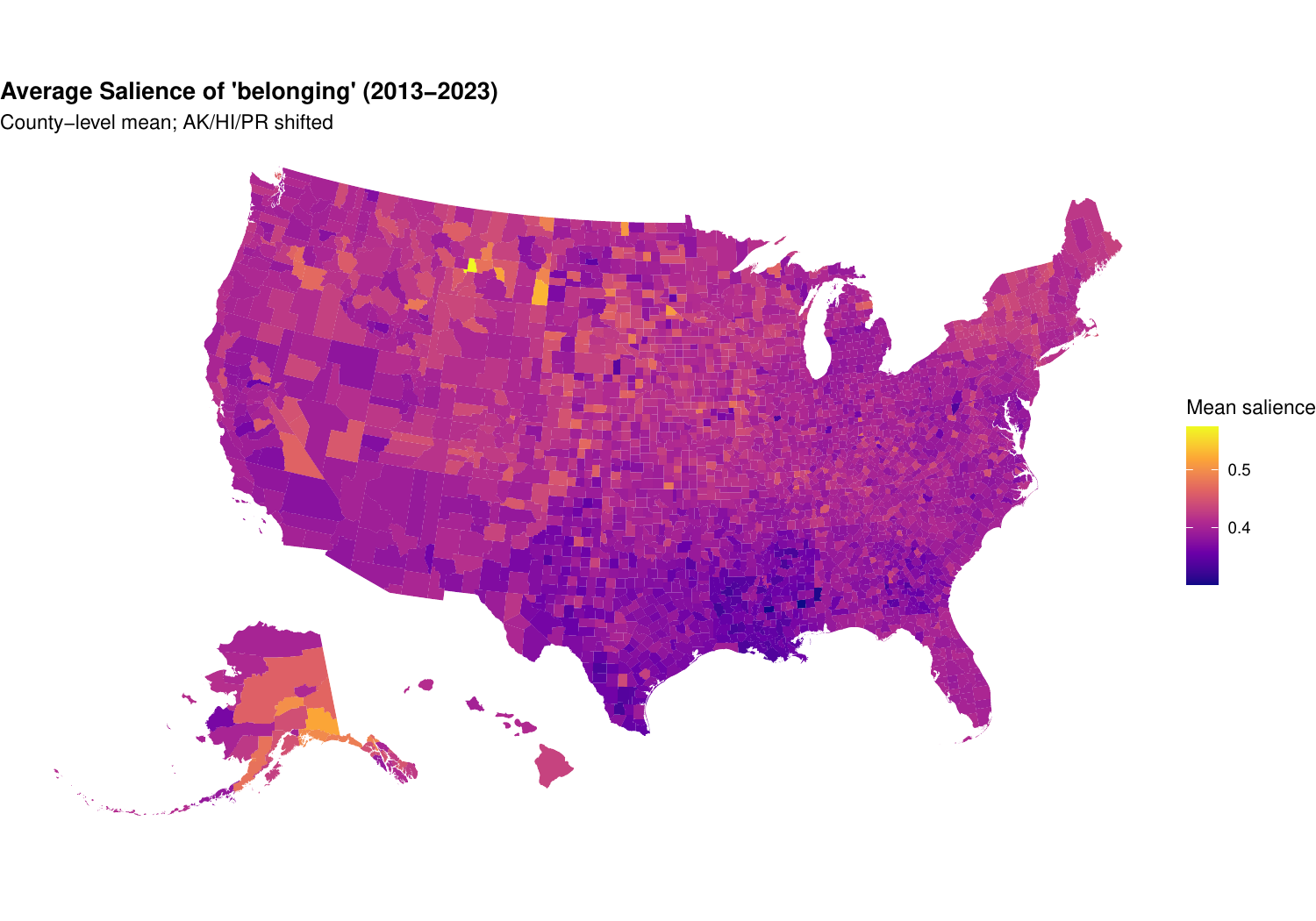}
\includegraphics[width=0.4\textwidth]{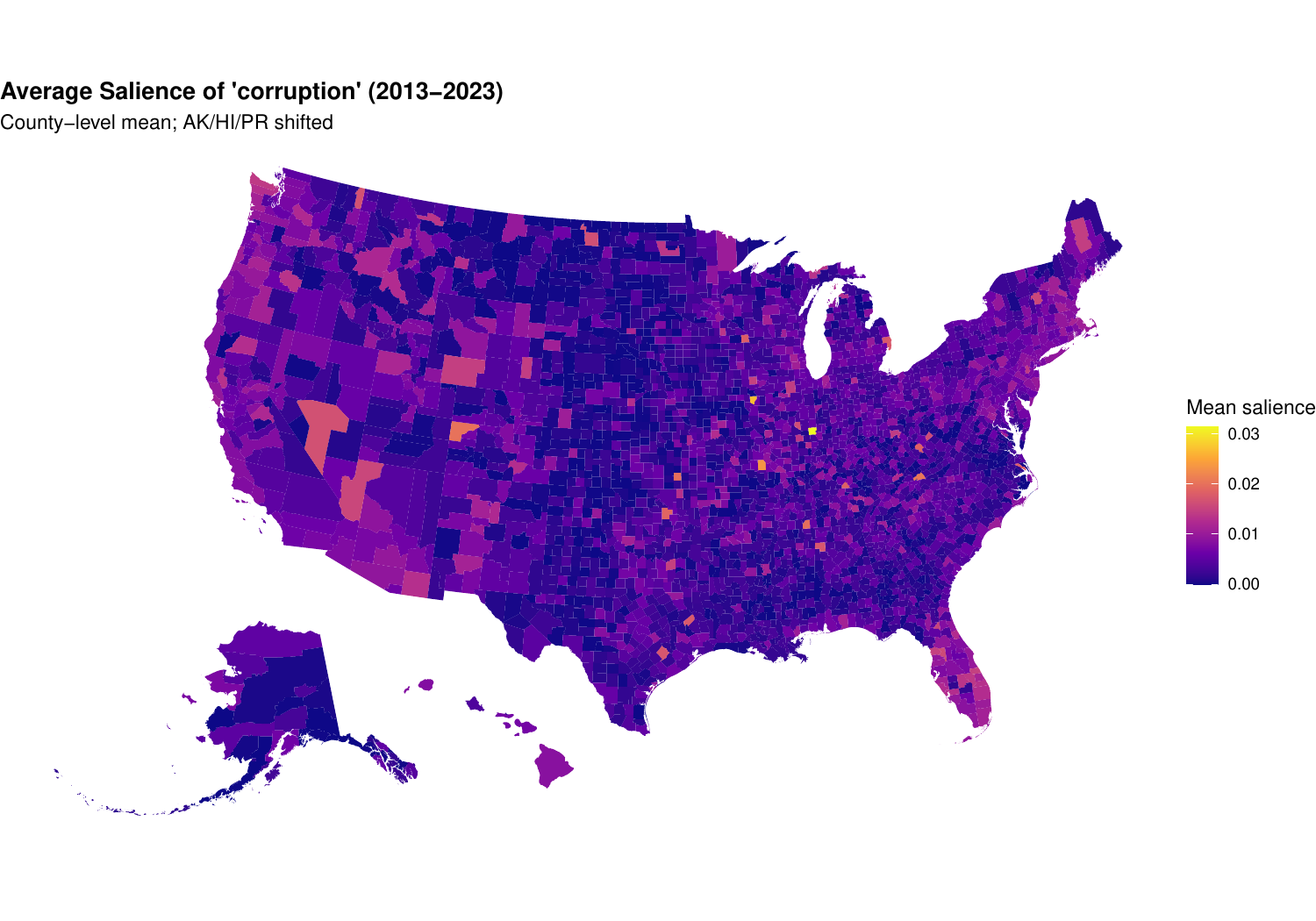} \\
\includegraphics[width=0.4\textwidth]{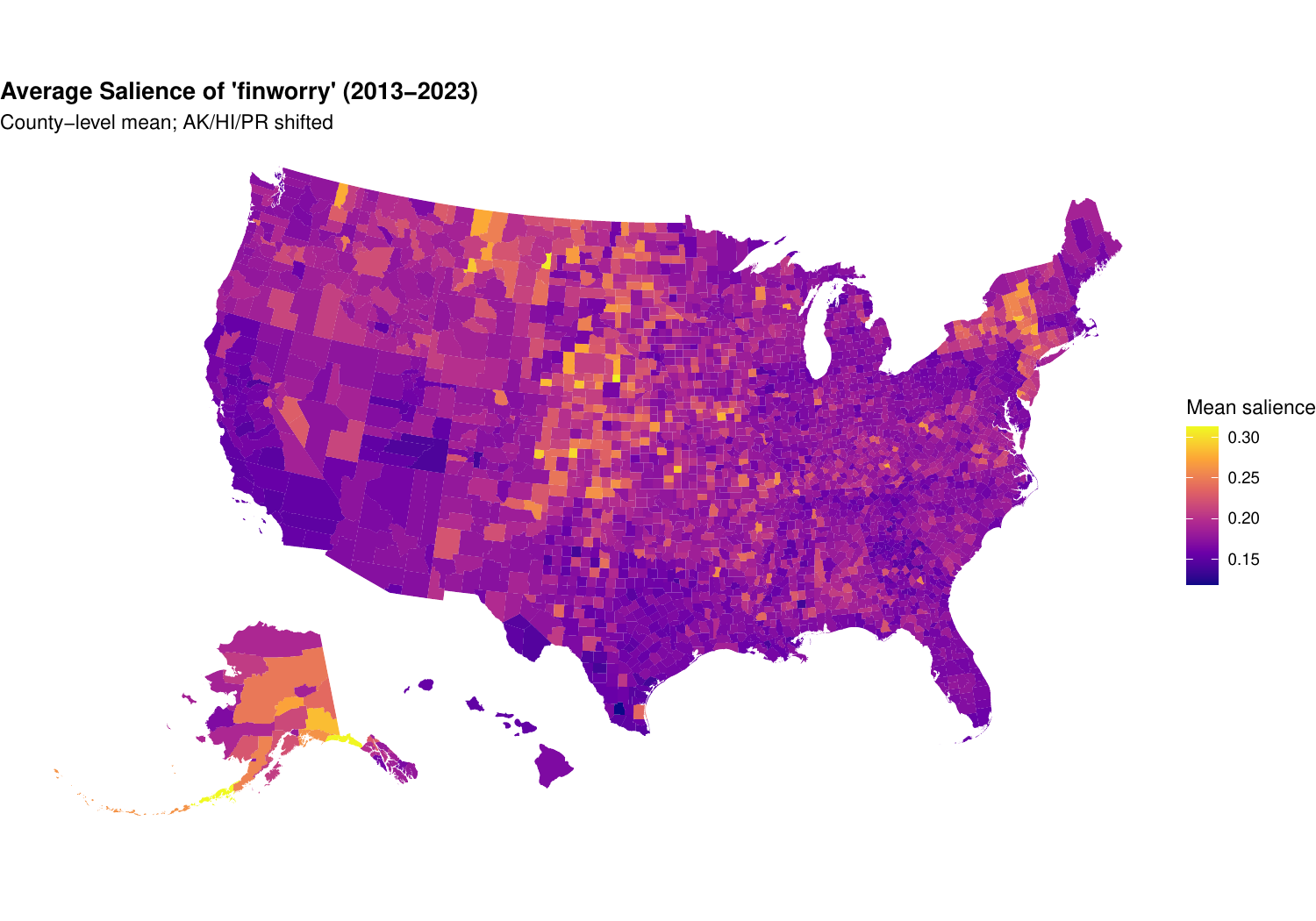}
\includegraphics[width=0.4\textwidth]{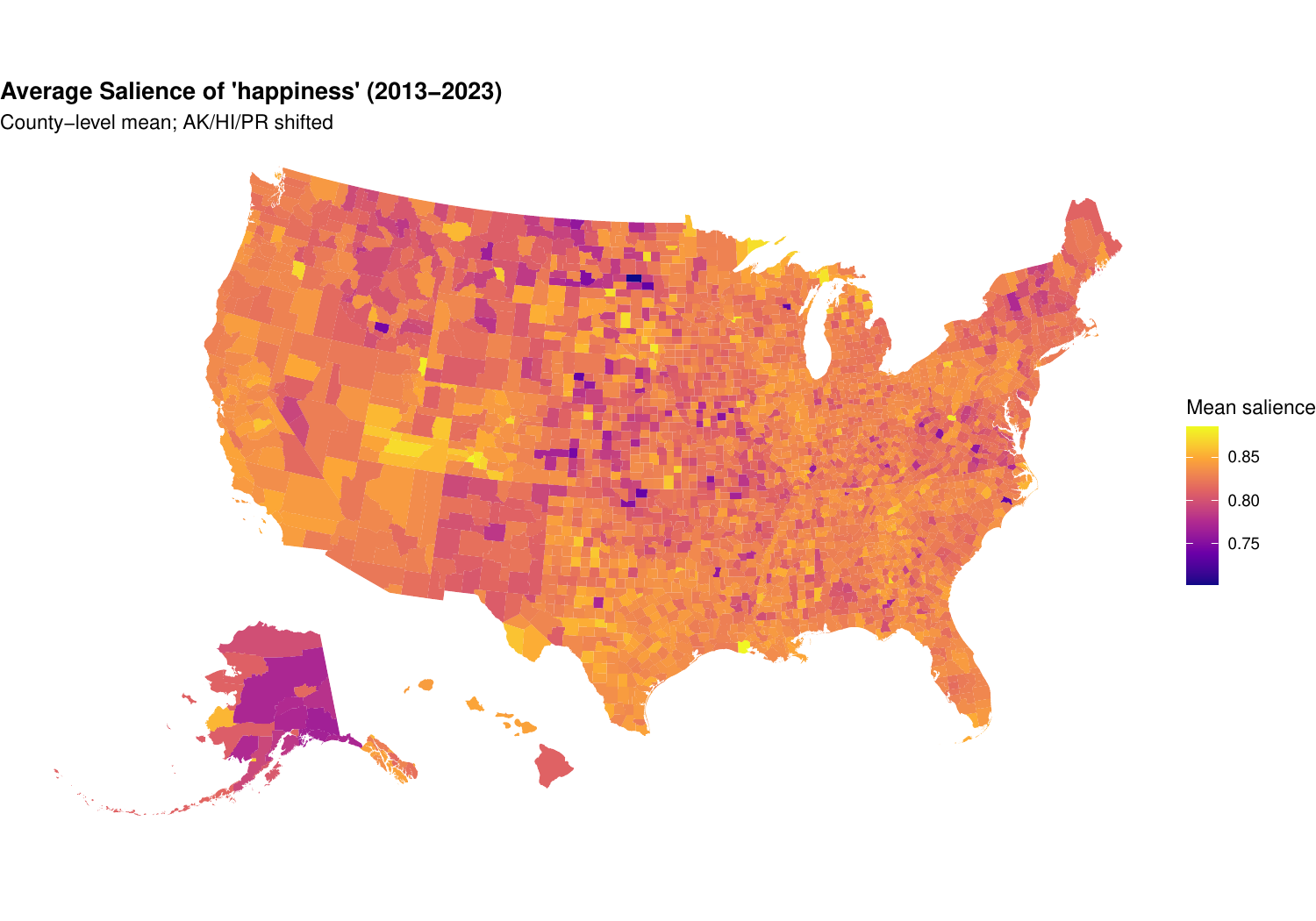}
\caption{Spatial variation in the average salience (= share of tweets expressing the dimension) of selected flourishing dimensions across U.S.\ counties (2013–2023).  
Each panel shows the county-level mean salience—the share of tweets expressing a given dimension—computed over the full 2013–2023 period.  
Dimensions are selected to illustrate both highly prevalent (\texttt{happiness}) and less frequent but thematically distinct domains (\texttt{belonging}, \texttt{finworry}, \texttt{corruption}).  
Color scales are independent across panels to emphasize within-dimension spatial contrasts.  
Geographic patterns reveal coherent regional structure even for low-base topics, indicating that public expressions of well-being, material security, and civic concern vary systematically across the United States.  
Alaska, Hawaii, and Puerto Rico are shown in compact shifted layout.}
\label{fig:salience_by_geo}
\end{figure}

Salience also varies systematically across U.S.\ counties. Figure~\ref{fig:salience_by_geo} illustrates four representative dimensions chosen to span highly salient (\texttt{happiness}) and more specialized domains (\texttt{belonging}, \texttt{finworry}, \texttt{corruption}). Each panel uses an independent color scale to preserve within-dimension contrast given large differences in absolute salience across topics; thus, colors are not comparable across panels in absolute terms. Three observations emerge.

\paragraph{Wide baseline with regional texture (happiness).}
The \texttt{happiness} panel shows that positive affect is a pervasive topic of online discourse nationwide, with broad coverage across most counties. Despite this high baseline, spatial texture is evident: contiguous bands of higher salience appear across large swaths of the interior West and the central corridor, while some metropolitan belts and parts of the coastal Northeast show relatively lower salience within this panel’s range. This pattern is consistent with a ubiquitous affective register overlaid by regional gradients in public expression.

\paragraph{Social and economic anchoring (belonging, finworry).}
Both \texttt{belonging} and \texttt{finworry} exhibit marked geographic structure, though in distinct ways. The \texttt{belonging} panel shows a broad band of moderate-to-high salience across the Midwest and interior regions, with comparatively lower values around several major metropolitan areas, indicating that discourse on community and social rootedness is not uniformly distributed but reflects regional variation in collective identity. In contrast, the \texttt{finworry} dimension highlights counties where economic concern and material insecurity occupy a more visible place in public discourse. Elevated salience is most pronounced across the Great Plains and portions of the rural Midwest—areas characterized by agricultural economies and relatively low population density, where financial concerns may reflect exposure to market fluctuations and farm income volatility rather than persistent unemployment or structural labor market weakness\cite{partridge2005persistent, weber2017ruralpoverty, mather2018usregionalinequality}.  Together, these two dimensions suggest that both social anchoring and financial anxiety form coherent spatial patterns that parallel long-standing socio-economic and demographic divides rather than random noise.

\paragraph{Low-base, event-sensitive discourse (corruption).}
The \texttt{corruption} panel has a low national base rate, as expected for a specialized civic topic. Against that low baseline, the map reveals localized clusters and corridors of higher salience, including in parts of the South and selected urban counties elsewhere. Given the topic’s episodic nature, such clusters may reflect media cycles, state-level political dynamics, or local scandals that periodically elevate discussion \citep[e.g.,][]{BauhrGrimes2014_Exposure,CostasPerez2012_CorruptionMedia,pew2014_polarization_media_habits,entman2007framing,Hayes2025,stockemer2019corruptionmedia}. The spatial heterogeneity observed in corruption-related discourse suggests, but does not definitively establish, that even low-salience civic themes may leave geographically patterned traces in public discussion. Further analysis would be required to determine whether these patterns reflect coherent regional structures or stochastic variation.

Taken together, these panels demonstrate that county-level salience surfaces display spatial structure across domains that differ widely in baseline prevalence.
While salience should not be read as prevalence of private attitudes, its spatial organization provides actionable signal for research. In particular, the maps can be leveraged to:
\begin{itemize}
  \item identify regional contrasts and contiguous ``corridors'' of discourse intensity for specific flourishing domains;
  \item generate hypotheses for how local institutions, media markets, or socio-demographic composition modulate the public expression of well-being;
  \item inform spatial models that incorporate salience as an \emph{opportunity} or \emph{visibility} weight when comparing indicator levels across places.
\end{itemize}
Methodologically, county-level salience layers can be paired with classical spatial statistics and covariates such as urbanicity, broadband access, religious adherence, or civic participation. Because salience is constructed as a share over the local tweet universe, differences across space reflect both the frequency with which a theme is expressed and the opportunity to express it in that information ecosystem. As such, the maps highlight that spatially explicit indicators can complement national aggregates: they reveal where discourse about flourishing concentrates, how it varies across domains, and where changes over time should be monitored most closely.

\subsection{The Geography of Religious Discourse and the ``Bible Belt''}

To contextualize our online indicator \texttt{believegod}, we compare it with offline measures  of religiosity derived from the 2020 \emph{Religious Congregations and Membership Study} (RCMS), made available through the \emph{Association of Religion Data Archives} (ARDA) \citep{arda2023}. 
The RCMS dataset provides county-level counts of religious \emph{adherents} and \emph{congregations} across all denominational families in the United States. 
In this context, \textit{adherence} refers to institutional membership or affiliation, i.e., individuals formally associated with a congregation, not to personal belief or practice. 
From these data, we compute two indicators: (1) the total share of county population affiliated with any religious denomination (\emph{All adherents}), and (2) the share specifically belonging to evangelical Protestant groups (\emph{Evangelical adherents}), following the denominational classification used in the RCMS.

Because these three variables are expressed on different scales (population shares versus relative frequencies of social media discourse), we standardize them within the set of U.S. counties to z-scores. 
This normalization allows for meaningful visual comparison of their spatial patterns while preserving each variable's internal variation. 
Higher standardized values thus represent counties where religious adherence or online religious discourse is above the national average.

\begin{figure}[ht!]
\centering
\includegraphics[width=0.95\textwidth]{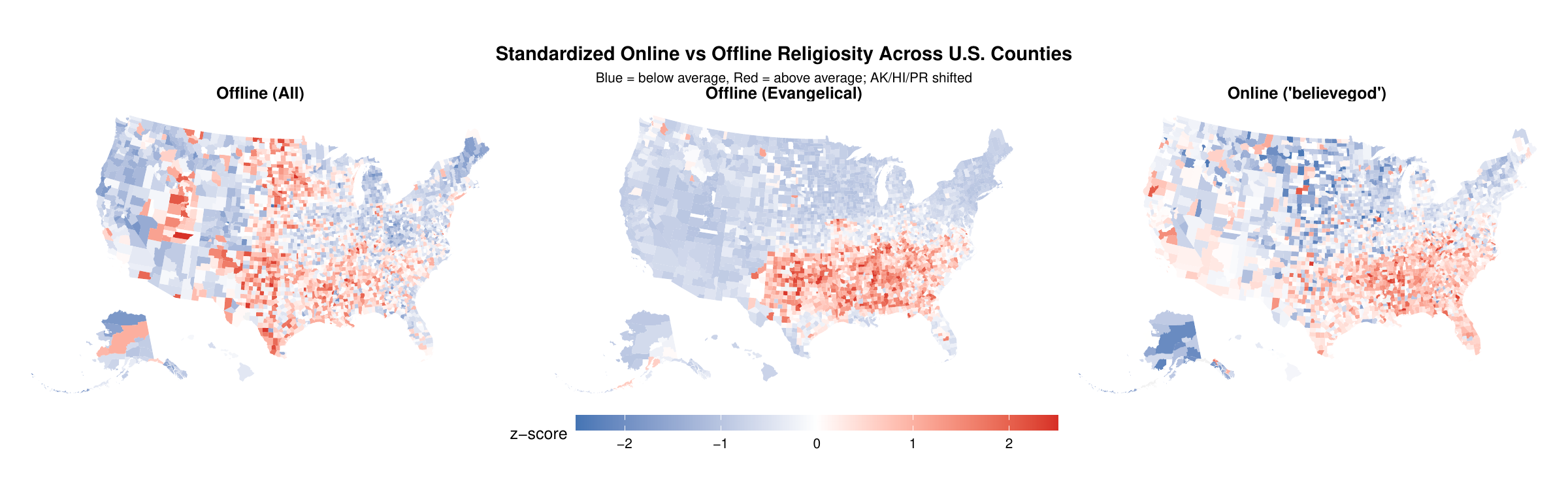}
\caption{Standardized offline and online religiosity across U.S. counties. 
Blue areas indicate below-average values, red areas above-average values (z-scores). 
The left and right panels show total and evangelical religious adherence from the RCMS 2020 (ARDA), 
while the central panel displays the standardized salience of religious discourse on social media (\texttt{believegod}, 2013–2023).}
\label{fig:biblebelt}
\end{figure}

As shown in Figure~\ref{fig:biblebelt}, the spatial distributions of these standardized indicators reveal a clear, though not exclusive, correspondence of the online indicator with the so-called \emph{Bible Belt}, historically defined as the region characterized by strong evangelical presence and conservative Protestant traditions \citep{Heatwole01021978,silk1988biblebelt,Tweedie78}. 
Both offline indicators show concentrated religious adherence in the southeastern states and parts of the lower Midwest, while the online \texttt{believegod} dimension exhibits a closely aligned yet somewhat broader pattern extending into the central Plains and Midwest.

It is important to emphasize that our social media indicator does not measure individual faith or theological belief in God, but rather the \emph{public salience} of religious language in digital communication, that is, how often people \emph{talk about} religion in the online public sphere. 
This distinction clarifies the conceptual parallel among the three measures: while offline adherence captures institutional embeddedness, the online dimension reflects the cultural visibility of religion in everyday discourse. 
Despite their different nature and time span (the RCMS reflects institutional membership in 2020, whereas the online data aggregate activity from 2013 to 2023) the high degree of spatial overlap suggests that the geography of religiosity in the United States is reflected in both institutional and online measures.

\begin{table}

\caption{Pairwise correlations among standardized religiosity measures (z-scores). Pearson and Spearman coefficients for county- and state-level data.}
\label{tab:religiosity-correlations}
\centering
\begin{tabular}[t]{llrrr}
\toprule
Level & Pair & Pearson & Spearman & N\\
\midrule
County & Online vs Evangelical & \textbf{0.385} & \textbf{0.507} & 3099\\
County & Online vs All & 0.099 & 0.158 & 3099\\
County & Evangelical vs All & 0.596 & 0.385 & 3099\\
\addlinespace
State & Online vs Evangelical & \textbf{0.760} & \textbf{0.660} & 51\\
State & Online vs All & 0.351 & 0.318 & 51\\
State & Evangelical vs All & 0.457 & 0.313 & 51\\
\bottomrule
\end{tabular}
\end{table}
To quantify the strength of these spatial correspondences, Table~\ref{tab:religiosity-correlations} reports the pairwise correlations among the standardized indicators of online and offline religiosity. 
At the county level, the correlation between the online \texttt{believegod} dimension and evangelical adherence is moderate (\(r = 0.39\); \(\rho = 0.51\)), whereas its association with overall religious adherence is weak (\(r = 0.10\); \(\rho = 0.16\)). 
The two institutional measures---total and evangelical adherence---remain positively related (\(r = 0.60\)), but the difference in magnitude suggests that the online indicator aligns more closely with the evangelical component of U.S. religiosity than with the aggregate pattern of church membership. 
When aggregated to the state level, the correlation between online and evangelical religiosity becomes very strong (\(r = 0.76\); \(\rho = 0.66\)), indicating that regional differences are highly consistent once local variation is averaged out. 
By contrast, the correspondence between online discourse and overall adherence remains modest (\(r = 0.35\)), and the institutional correlation between evangelical and total adherence weakens to \(r = 0.46\), reflecting heterogeneity in denominational composition across states.

Overall, these results show that social media discourse about religion mirrors the geography of evangelical adherence more than that of general church affiliation. 
This reinforces our conceptual interpretation: the \texttt{believegod} dimension captures the \emph{public salience} of religious language rather than personal faith or institutional belonging. 
In other words, while digital discussions of religion are concentrated in the same broad regions that constitute the historical Bible Belt, their strength and diffusion reveal a communicative rather than purely confessional geography of religiosity in the contemporary United States.

\subsection{Rural–Urban Differences in Expressed Well-Being}

To investigate how flourishing-related expressions vary across different settlement contexts, we estimated county-level regressions of the form:

\begin{equation}
y_i = \alpha + \beta_1 \texttt{Rural}_i + \beta_2 \log(\texttt{ntweets}_i) + \beta_3 \log(\texttt{population}_i) + \varepsilon_i,
\end{equation}
where $y_i$ is the average score of a given well-being dimension in county $i$, $\texttt{ntweets}_i$ and $\texttt{population}_i$ are controls for social media activity and population size, respectively.
The variable $\texttt{Rural}_i$ is constructed as a binary indicator based on the USDA Rural–Urban Continuum Codes \cite{USDA_RUCC_2024}, 
taking the value $\texttt{Rural}_i = 0$ for metropolitan counties (RUCC codes $1$–$3$) and $\texttt{Rural}_i = 1$ for nonmetropolitan counties (RUCC codes $4$–$9$), which represent progressively smaller and more remote communities.
Separate models were estimated for each of the 46 flourishing indicators.

\begin{figure}[ht!]
\centering
\includegraphics[width=0.9\textwidth]{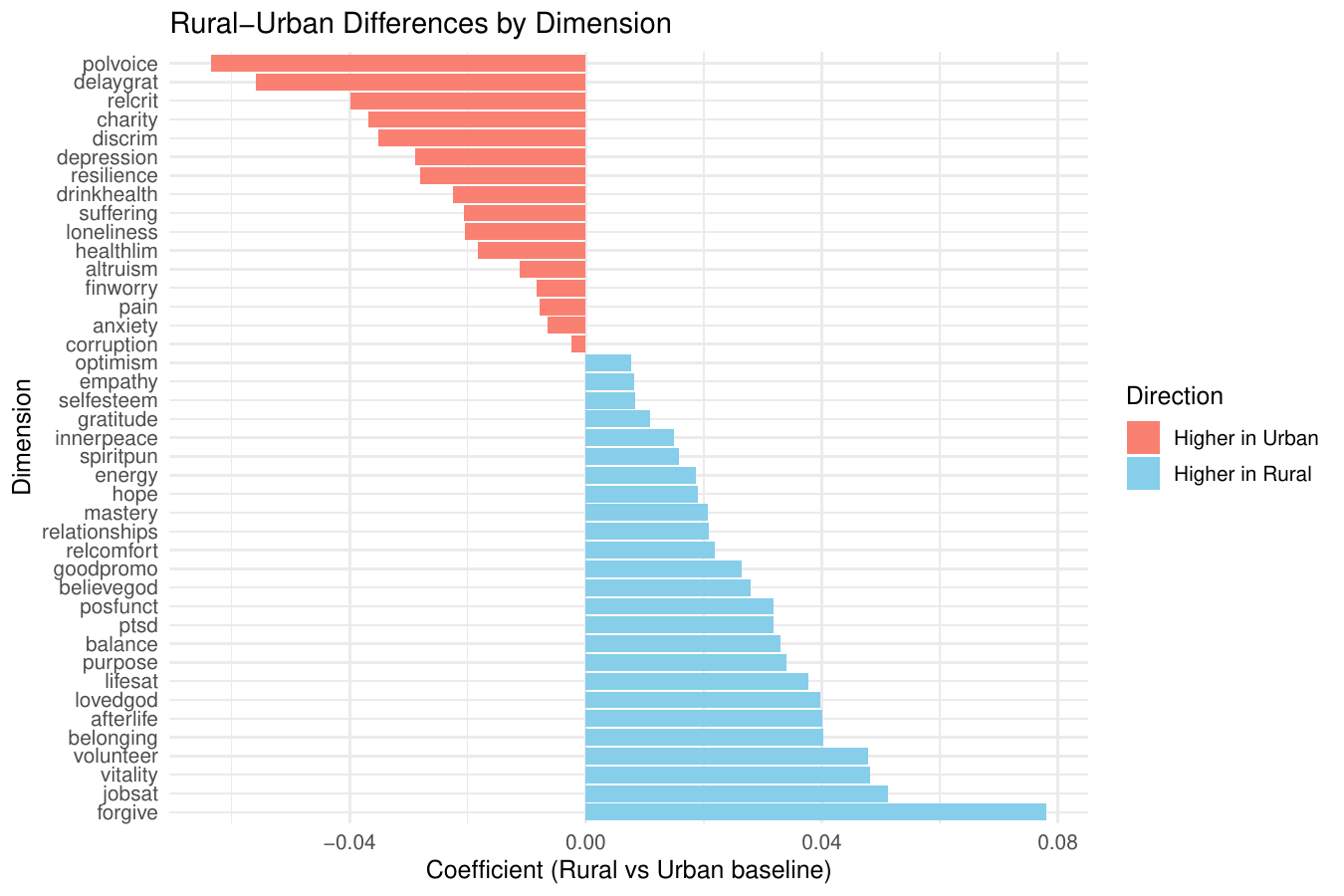}
\caption{Estimated differences in well-being dimensions between rural and urban U.S.\ counties (2013–2023). 
Each bar represents the estimated coefficient $\beta_1$ from separate linear models of the form 
$y_i = \alpha + \beta_1 \text{Rural}_i + \beta_2 \log(\text{ntweets}_i) + \beta_3 \log(\text{population}_i) + \varepsilon_i$, 
where $y_i$ is the county-level mean value of each flourishing indicator. 
Positive coefficients indicate higher levels in rural counties, negative coefficients higher levels in urban counties.
Bars are shown only for statistically significant effects (unadjusted $p<0.05$). 
Results highlight that expressive dimensions of religious belief, purpose, and subjective well-being tend to be stronger in rural areas, 
whereas civic and moral concern dimensions (e.g., \textit{political voice}, \textit{delayed gratification}, \textit{charity}) 
are more salient in urban contexts.}
\label{fig:ruralurban}
\end{figure}

Figure~\ref{fig:ruralurban} summarizes the estimated $\beta_1$ coefficients, 
showing only statistically significant contrasts (unadjusted  $p$-values, $p < 0.05$).
Positive values indicate higher average expression in rural counties relative to urban ones, and vice versa.

The pattern is strikingly consistent with classical sociological and psychological distinctions between urban and rural environments \citep[e.g.,][]{wirth1938urbanism,tuan1975topophilia,fischer1975subcultural,stewart1958community,sampson2012great}.
Indicators tied to \emph{religious faith and moral virtue} (\texttt{believegod}, \texttt{relcomfort}, \texttt{forgive}, \texttt{lovedgod}), 
as well as \emph{happiness and purpose} (\texttt{lifesat}, \texttt{purpose}, \texttt{balance}, \texttt{volunteer}), 
show higher prevalence in rural counties.
In contrast, urban counties exhibit stronger expression of \emph{civic engagement and critical reflection} dimensions (notably \texttt{polvoice}, \textit{delayed gratification}, \texttt{charity}, and \texttt{relcrit}) which correspond to forms of public or moral discourse more common in dense, pluralistic social environments.

These associations are correlational, not causal, and partly reflect compositional and contextual differences in the underlying population and online discourse volume. 
Nonetheless, the patterns observed indicate that spatially resolved, language-based well-being indicators are sensitive to geographic variation in the relative salience of human flourishing dimensions across the rural–urban continuum, even after accounting for communication volume and population size.

\clearpage 

\section{Technical Validation}

\subsection{Fine-Tuning Setup}
The LLM was fine-tuned to answer questions of the type presented below. We used a system instruction and a user prompt that requested classification across a fixed set of well-being dimensions, returning a compact JSON dictionary containing only the applicable dimensions rated on a {low, medium, high} scale. The prompt was structured as follows:
\begin{quote}
\small
\textbf{System:} \\
You are a medical professional expert in subjective well-being. Answer the question truthfully. \\

\textbf{User:} \\
Please classify the following text based on the well-being dimensions listed below. Use only the scale: \texttt{low}, \texttt{medium}, and \texttt{high}. Return a JSON dictionary that contains only the well-being dimensions that apply. Do not explain your reasoning.

\medskip
\noindent
\textbf{Well-being dimensions:}

\small
\texttt{[ C1 = `happiness', C2 = `resilience', C3 = `self-esteem', C4 = `life satisfaction', C5 = `fear of future', C6 = `vitality', C7 = `having energy', C8 = `positive functioning', C9 = `expressing job satisfaction', C10 = `expressing optimism', C11 = `peace with thoughts and feelings', C12 = `purpose in life', C13 = `depression', C14 = `anxiety', C15 = `suffering', C16 = `feeling pain', C17 = `expressing altruism', C18 = `loneliness', C19 = `quality of relationships', C20 = `belonging to society', C21 = `expressing gratitude', C22 = `expressing trust', C23 = `feeling trusted', C24 = `balance in the various aspects of own life', C25 = `mastery (ability or capability)', C26 = `perceiving discrimination', C27 = `feeling loved by god', C28 = `belief in god', C29 = `religious criticism', C30 = `spiritual punishment', C31 = `feeling religious comfort', C32 = `financial or material worry', C33 = `life after death belief', C34 = `volunteering', C35 = `charitable giving or helping', C36 = `seeking for forgiveness', C37 = `feeling having a political voice', C38 = `expressing government approval', C39 = `having hope', C40 = `promoting good', C41 = `expressing delayed gratification', C42 = `ptsd (post-traumatic stress disorder)', C43 = `describing smoking related health issues', C44 = `describing drinking related health issues', C45 = `describing health limitations', C46 = `expressing empathy' ]}.

\medskip
Here is the text: \\
\textit{"your tweet here"}.

\medskip
\textbf{Assistant:} \\
Answer: \\
\end{quote}
During fine-tuning, the model was given examples such as the tweet \textit{"not going to be able to sleep for the rest of the week dude"}, with the corresponding completion of the prompt shown below:
\begin{quote}
Answer: \\
\texttt{\{"C1": "low", "C6": "low", "C10": "low", "C4": "low", "C20": "low", "C39": "low", "C45": "medium"\}}
\end{quote}

It is key to remark that the model has been asked to answer only about the dimensions that apply and together with imposing the compact answer \texttt{\{"CX": "YYY"\}}, helped us to both quickly parse the results as well as saving output tokens, resulting in a faster inference.

\subsubsection{Prompt on attitudes towards migration}
In this case the prompt is a simple one.
\begin{quote}
\small
\textbf{System:} \\
You are an expert on migration. Answer the question truthfully. \\

\textbf{User:} \\
The next text is a tweet probably about migration:\\

"your tweet here"\\

Analyze carefully the tweet and assign it to the most relevant category among those listed below. Do not explain your answer and return only a number.\\

Category numbers: 1 = 'pro-immigration'; 2 = 'anti-immigration'; 3 = 'neutral'; 4 = 'unrelated'.

\medskip
\textbf{Assistant:} \\
Answer: \\
\end{quote}
and the \texttt{migmood} indicator is built assigning $+1$ to `pro-immigration'; $-1$ to  `anti-immigration' and $0$ to  `neutral'.

\subsubsection{Prompt of perception of corruption}
This prompt is more complex that the one about attitudes towards migration because this research implies deeper analysis of this topic.
\begin{quote}
\small
\textbf{System:} \\
You are an expert on corruption issues. Answer the question truthfully. \\

\textbf{User:} \\
I\'m studying perception of corruption. Here is a tweet:\\

"your tweet here"\\

I have five questions you need to answer.\\
Q1: Is the tweet about corruption?  1 = yes, 2 = no.\\

Q2: Does the tweet: 1 = express distrust in governmental or judicial institutions; 2 = mention specific leaders or organizations to be blamed; 3 = it is not about corruption. Answer with a number. Do not give a textual explanation.\\

Q3: Does the tweet: 1 = describe personal experiences with corruption; 2 = reflect broader societal issues, such as inequality or governance problems; 3 = it is not about corruption. Answer with a number. Do not give a textual explanation.\\

Q4: Does the tweet: report any specific type of corruption? 1 = generic corruption, 2 = Bribery, 3 = Fraud, 4 = Nepotism, 5 = Abuse of power, 6 = Conflict of interest, 7 = Money laundering, 8 = Tax evasion, 9 = Insider trading, 10 = Corporate fraud, 11 = Police corruption, 12 = Judicial corruption, 13 = Extortion, 14 = Cover-up,  15 = Whistleblower, 16 = not about corruption. Answer with a number or a sequence of numbers separated by commas. Do not give a textual explanation. \\

Q5: Are emotions such as anger, frustration, or distrust prevalent? 1 = yes, 2 = no. Answer with a number. Do not give a textual explanation.\\

Q6: Does the tweets reflect narratives shaped by news outlets? 1 = yes, 2 = no.\\

Use the following template: \\
Q1 = [your answer];\\
Q2 = [your answer];\\
Q3 = [your answer];\\
Q4 = [your answer];\\
Q5 = [your answer];\\
Q6 = [your answer].\\
\\
\textbf{Assistant:} \\
Answer: \\
\end{quote}
and the measure \texttt{corruption} is built solely on Q1. The dataset reports also answers to the other questions in the form of binary variables so that, e.g.,  \texttt{q4\_1 = 1} means in question Q4, the first option has been selected.

\subsection{Model Performance on the Training and Validation Sets}

Starting from a set of 10,000 randomly selected tweets, we distilled a final corpus of 4,581 human-supervised and coded tweets, which was used for fine-tuning.  
This dataset was generated following the approach introduced in \cite{finetuning2024}. First, a range of diverse LLMs — from small to large Llama models up to ChatGPT-4 — were given all tweets to classify using prompts similar (and slightly adjusted for each model) to the one presented above.  
This procedure produced multiple sets of 10,000 unsupervised machine classifications.  

Next, a team of 12 human coders was randomly presented with each tweet, together with all categories and ratings generated by all models.  
The coders were instructed to reject incorrect classifications and ratings that did not apply.  
We then split the resulting data into training and test sets of 2,404 and 2,177 tweets, respectively.

Because each tweet could be classified along 46 well-being dimensions simultaneously, the total number of coding instances to analyze was 110,584 and 100,142 for the training and test sets, respectively.  
In practice, each tweet is replicated according to the number of applicable flourishing dimensions.  
In our application, it is crucial not only to correctly identify the dimensions that apply but also to discard those that do not. This is not a symmetric problem.  

Indeed, we observed that non–fine-tuned models exhibit distinct biases: larger models (including ChatGPT-4) tend to classify tweets along too few dimensions, leading to many false negatives, whereas smaller models classify along too many, resulting in many false positives.  
For this reason, in addition to accuracy ($Acc$, higher is better), we also report the Jaccard Index ($J$, higher is better) and the Hamming Loss ($H$, lower is better), which evaluate correctness across the entire string of true and false labels.  

Table~\ref{tab:validation} reports the validation results.  
For comparison, we also include a Llama~2 model with 7B parameters — more than twice the size of the model used here.  
The results show that small models perform poorly if not fine-tuned but, once fine-tuned, their performance approaches that of ChatGPT-4, despite the latter having 587 times more parameters.  
We opted for Llama~3.2~3B rather than Llama~2~7B for both computational efficiency and multilingual scalability, as the terms of use for Llama~2 restrict usage to English-language applications.  Notice that, by chance, the initial validation set of 2,177 tweets appears harder to classify also by ChatGPT~4 and Llama~2~7B as their accuracy goes down by about 10 points both in terms of Jaccard metrics and standard accuracy. 

In Table~\ref{tab:validation} we also added the analysis of a growing training set moving from 52\% to 80\%, further to 98\% of the whole set of labeled data. As it can be seen, Llama~3.2~3B seems stable as the size of the training data increases. Moreover, the performance remains the same between the validation (out-sample) and training (in-sample) data.

\begin{table}[ht!]
\centering
\caption{Validation results comparing model size, fine-tuning, and performance metrics. The model in bold is the one chosen for scaling the analysis to the 2.6 billion dataset. This model used 98\% of the whole labelled dataset during fine-tuning. The 2\% has been left out just for sanity checking.}
\label{tab:validation}
\begin{tabular}{lrlccccc}
\hline
\textbf{Model} & \textbf{Param} & \textbf{Validation} & \textbf{Set} & \textbf{$\%$ labeled data}& \textbf{Acc} & $\mathbf{J}$ & $\mathbf{H}$ \\
\hline
Llama~3.2 & 3B & naive & in-sample & 52 &0.29 & 0.28 & 0.71 \\
Llama~2 & 7B & naive & in-sample & 52 &0.34 & 0.31 & 0.68 \\
Llama~3.2 & 3B & fine-tuned & in-sample & 52 &0.79 & 0.77 & 0.22 \\
Llama~3.2 & 3B & fine-tuned & in-sample & 80 & 0.87 & 0.78   & 0.12 \\
\textbf{Llama~3.2} & \textbf{3B} & \textbf{fine-tuned} & \textbf{in-sample} & \textbf{98} & \textbf{0.87} & \textbf{0.78}   & \textbf{0.13} \\
Llama~2 & 7B & fine-tuned & in-sample &  52 &0.85 & 0.93 & 0.06 \\
ChatGPT-4 & 1,760B & naive & in-sample & 52 & 0.94 & 0.93 & 0.05 \\
\hline
Llama~3.2 & 3B & naive & out-sample & 48&0.29 & 0.29 & 0.71 \\
Llama~2 & 7B & naive & out-sample  & 48&0.34 & 0.33 & 0.66 \\
Llama~3.2 & 3B & fine-tuned & out-sample & 48& 0.76 & 0.74 & 0.23 \\
Llama~3.2 & 3B & fine-tuned & out-sample & 20 & 0.85 & 0.75   & 0.13 \\
\textbf{Llama~3.2} & \textbf{3B} & \textbf{fine-tuned} & \textbf{out-sample} & \textbf{2} & \textbf{0.88} & \textbf{0.79}   & \textbf{0.12} \\
Llama~2 & 7B & fine-tuned & out-sample  &48& 0.85 & 0.84 & 0.15 \\
ChatGPT-4 & 1,760B & naive & out-sample  & 48&0.86 & 0.84 & 0.13 \\
\hline
\end{tabular}
\end{table}

\clearpage

\section{Validation with External Data Sources and Other Sentiment Indicators}

Unlike traditional sentiment indices that rely on fixed lexicons or polarity dictionaries, the HFGI employs context-aware large language models capable of interpreting meaning beyond individual word choice. This design ensures semantic continuity across linguistic change, slang, and platform evolution over the decade-long study period. 

In this section, we present a set of straightforward association and convergence analyses, without any causal ambition, but only to illustrate how the HFGI can complement survey-based and official statistics in advancing the empirical study of human flourishing.

\subsection{Convergent validity with an independent sentiment index (TSGI)}

To examine how closely the tweet-based happiness indicator aligns with the external sentiment measure, we consider the  Twitter Sentiment Geographical Index (TSGI) \cite{chai2023tsgi_scidata}. This sentiment score is evaluated on a $[0,1]$ scale and is based on specialized fine-tuned BERT classifier. We aggregated the TSGI sentiment score at monthly level to compare it with our HFGI of \texttt{happiness}. The latter varies in $[-1,+1]$. We expect these indicators to behave similarly but not to be exactly the same. These indicators are calculate on the same exact raw Twitter data.

To make sense of their different scales
we estimated a simple linear model after standardizing both variables to zero mean and unit variance:
\[
z_{\text{happiness},i} = \beta_0 + \beta_1 \, z_{\text{sentiment},i} + \varepsilon_i.
\]
The results reveal a highly significant positive association between the two indicators ($\beta_1 = 0.62$, $\text{SE} = 0.0013$, $t = 487$, $p < 0.001$), with an intercept not significantly different from zero ($\beta_0 \approx 0$). The model explains approximately $R^{2} = 0.39$ of the variance in standardized happiness scores, corresponding to a Pearson correlation of $r = 0.62$ between the two measures. 

This relatively high correlation indicates that the two independently derived indicators capture a common underlying affective signal: regions and periods characterized by more positive linguistic sentiment tend also to display higher levels of expressed happiness. However, the relationship is not one-to-one, and roughly 60\% of the variance in happiness remains unexplained by the sentiment measure. This residual variation suggests that, while both indicators reflect general affective tone in social media content, they emphasize partially distinct constructs, happiness being a more evaluative or self-referential dimension of well-being, whereas sentiment captures broader lexical positivity. The two indicators should therefore be regarded as complementary rather than interchangeable measures of collective mood.

\subsection{External validation with CDC mental health data}

To assess the validity of our tweet-based depression indicator, we compare it to
county-level estimates of \textit{frequent mental distress} (FMD) from the
CDC Behavioral Risk Factor Surveillance System (BRFSS) PLACES data
(2013–2023; \cite{CDC_PLACES_Methods,CDC_BRFSS_2013_overview}).
The FMD measure represents the percentage of adults reporting
at least fourteen days of poor mental health during the previous month.

\subsubsection{Observed patterns in survey and social-media data}

We first model the CDC data alone to characterize its spatial and temporal
structure:
\[
\text{FMD}_{c,t} = \alpha + \beta_{1}\,\text{rural}_{c} + \gamma_{t} + \varepsilon_{c,t},
\]
where $\text{rural}_{c}$ is the USDA Rural–Urban Continuum Code \cite{USDA_RUCC_2024} and
$\gamma_{t}$ are year fixed effects.  
Results show that FMD prevalence \emph{increases} significantly with rurality
($\hat{\beta}_{1}=0.39$, $p<10^{-16}$): rural counties report higher
levels of poor mental health than urban counties, consistent with known rural
disparities in mental-health burden
\cite{Peen2010_UrbanRuralPsych,Morales2020_RuralMH,Forrest2023_UrbanRural}.

Next, we estimate an analogous specification for our social-media–based depression
indicator:
\[
\text{HFGI}_{c,t} = \alpha + \beta_{1}\,\text{rural}_{c} + \gamma_{t} + \varepsilon_{c,t}.
\]
Here, the slope reverses sign ($\hat{\beta}_{1}=-0.00095$, $p<10^{-10}$),
indicating that more rural counties exhibit \emph{lower} levels of
depression-related discourse on Twitter.
At face value, this pattern contrasts with the CDC data.
However, such a reversal is predicted by established theories of public
expression and stigma:
rural contexts combine stronger social sanctions against emotional disclosure,
higher perceived stigma, and lower broadband penetration, all of which
reduce the public visibility of distress online
\cite{NoelleNeumann1974_Spiral,LinkPhelan2001_Stigma,MarwickBoyd2011_ContextCollapse,Pew2024_SocialMediaUse,NTIA2024_BroadbandFunding,Krumpal2013_SDB}.
Under these conditions, the underlying prevalence of mental distress may be high
(as in the CDC data), yet its \emph{expressed} signal in social media appears low.
Our indicator thus captures an \emph{expression propensity} rather than raw
prevalence.

\subsubsection{Reconciling survey and social-media signals}

To reconcile these perspectives, we estimate a joint model incorporating both
survey and behavioral data:
\[
\text{HFGI}_{c,t} = \alpha
+ \beta_{1}\,\text{FMD}_{c,t}
+ \beta_{2}\,\text{rural}_{c}
+ \beta_{3}\,(\text{FMD}_{c,t}\times\text{rural}_{c})
+ \gamma_{t}
+ \varepsilon_{c,t},
\]
weighted by the inverse variance of the county-level estimates.
The results confirm that our indicator increases with higher CDC distress
($\hat{\beta}_{1}=0.0023$, $p<10^{-16}$),
even after controlling for rurality and year.
The baseline rural effect remains negative
($\hat{\beta}_{2}=-0.0143$, $p<10^{-16}$),
consistent with lower overall online expression in rural areas,
while the positive interaction term
($\hat{\beta}_{3}=0.00063$, $p<10^{-16}$)
suggests that where rural distress is salient, its online manifestation becomes
more tightly coupled with the CDC prevalence.
The model explains over 90\% of variance ($R^{2}=0.91$),
primarily through year fixed effects that absorb national-level fluctuations.

\subsubsection{Theoretical synthesis}

These three models together demonstrate that our tweet-based depression indicator
captures \emph{expressed} mental distress rather than diagnostic prevalence.
Following communication-theory and stigma frameworks
\cite{NoelleNeumann1974_Spiral,LinkPhelan2001_Stigma,MarwickBoyd2011_ContextCollapse},
we conceptualize public distress expression as a behavioral function of both
true prevalence ($P_{c,t}$) and an \emph{expression propensity} ($E_{c,t}$)
shaped by local stigma and digital access:
\[
\mathrm{HFGI}_{c,t} = \theta_{0}
+ \theta_{1}\,P_{c,t}
+ \theta_{2}\,E_{c,t}
+ \theta_{3}\,P_{c,t}\!\times\!E_{c,t}
+ u_{c,t}.
\]
Rurality influences $E_{c,t}$ through stronger stigma and lower connectivity,
which suppress overt online disclosure.
Once this structural difference is controlled for, our depression indicator
aligns positively and significantly with CDC-reported mental distress,
demonstrating its construct validity while revealing systematic variation in
the \emph{visibility} of mental-health discourse across the U.S. social landscape.

\subsection{Validation with Transparency International Data}

To assess the construct validity of our tweet-based  indicator of perceived \texttt{corruption}, we compare it with the
\textit{Corruption Perceptions Index} (CPI) produced annually by Transparency International
\cite{TI_CPI_2023_report,TI_USA_CPI_2024,TI_CPI_Method_2025}.
The CPI provides a cross-national benchmark of perceived public-sector corruption on a scale from
0 (highly corrupt) to 100 (very clean). It is derived from multiple expert assessments and business
surveys and represents the most widely used international measure of perceived corruption.

Our indicator, by contrast, measures the \textit{relative prevalence of corruption-related expressions}
in social media discourse. Specifically, it captures the conditional mean of tweets classified under
the \texttt{corruption} dimension, aggregated annually at the state level and then averaged across
states to obtain a national series. The indicator takes values in $[0,1]$. Higher values indicate greater intensity or frequency of
corruption-related discussion. Because the CPI score increases (it is a transparency index in fact) as perceived corruption decreases,
a strong negative association between the two series would provide support for the validity of our
measure.

\begin{figure}[ht!]
\centering
\includegraphics[width=0.9\textwidth]{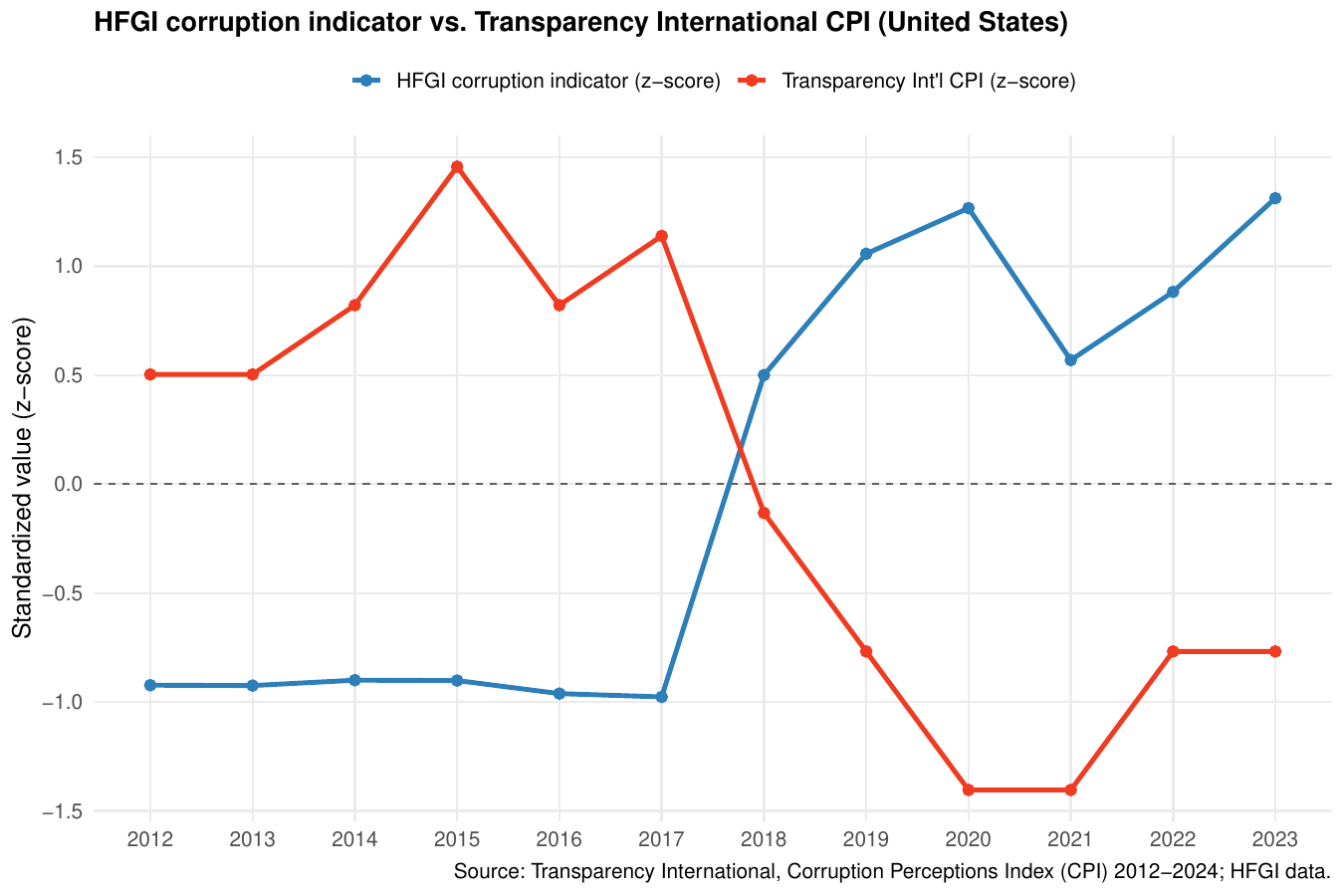}
\caption{Validation of the tweet-based perceived corruption indicator against the Transparency International
Corruption Perceptions Index (CPI) for the United States, 2012–2024 (truncated to 2023 because of the HGFI corruption indicator). Both series are standardized
(z-scores) to facilitate comparison. CPI: 0–100 scale where higher values indicate cleaner public
sectors. HFGI indicator: conditional mean in $[0,+1]$, where higher values indicate more
corruption-related expression. }
\label{fig:corruption_validation}
\end{figure}

Figure~\ref{fig:corruption_validation} compares the standardized (z-scored) annual series for the
United States from 2012 to 2024 (2023 for our indicator). The two measures display a clear inverse pattern: years of higher
CPI scores (cleaner public-sector perception) correspond to lower values of the tweet-based
corruption indicator, and vice versa. The Pearson correlation between the two series is
$r = -0.91$, indicating a very strong and statistically significant relationship.
To further quantify this association, we estimate a simple ordinary least squares model using standardized values of both variables:
\[
z_{\text{corruption}} = \alpha + \beta \, z_{\text{CPI}} + \varepsilon,
\]
where $z_{\text{corruption}}$ denotes the standardized tweet-based corruption indicator,
and $z_{\text{CPI}}$ is the standardized Corruption Perceptions Index score
(higher values indicating cleaner governance).
The estimated coefficients are:
\[
\hat{\alpha} = 0.14 \; (\mathrm{SE}=0.13), \qquad
\hat{\beta} = -1.01 \; (\mathrm{SE}=0.15), \qquad
R^{2} = 0.83, \; p < 0.001.
\]
The negative and highly significant slope confirms the expected inverse relationship:
years with higher CPI scores (cleaner perception) correspond to lower levels of corruption-related
discourse on social media.

The estimated slope of approximately $-1.0$ indicates that a one standard deviation increase in
CPI (a cleaner perception) is associated with a one standard deviation decrease in our corruption
indicator (more positive discourse regarding the absence of corruption). The model explains about
83\% of the yearly variation, further confirming the strong inverse association between the two
measures. 

Overall, this result demonstrates that the tweet-derived corruption dimension captures a signal
that moves coherently with a recognized international benchmark, supporting its external validity
as an indicator of corruption-related public sentiment.

\subsection{Associations between Climate Risk and Human Flourishing Dimensions}\label{sec:risk}

To explore how regional environmental exposure co-varies with different domains of human flourishing,
we computed pairwise Spearman correlations between county-level resilience-adjusted climate risk indicators
and social well-being dimensions inferred from social media data.
The eight climate risk variables (\texttt{Fire\_raj}, \texttt{Drought\_raj}, \texttt{Heat\_raj},
\texttt{Inland\_raj}, \texttt{Coastal\_raj}, \texttt{Wind\_raj}, \texttt{Mean\_raj}, and
\texttt{Overall\_raj\_pct}) are described in Table~\ref{tab:climate_risk_vars}.
Each represents a percentile score (0--100) from the \textit{Climate Risk and Resilience-adjusted Index}
(2005--2100) \cite{AlphaGeo_ClimateRisk2025}, where higher values indicate greater exposure to physical climate hazards after accounting
for local adaptation capacity.  At the county level, these indicators reflect the relative position of each
geographic unit within the global distribution of resilience-adjusted hazard intensity, integrating both
the frequency and potential severity of climate-related physical risks (e.g., drought, heat stress, flooding,
wind, wildfire) together with estimated adaptation and mitigation capacity.  In scientific terms, these
variables quantify the expected hazard exposure conditional on resilience-relevant features such as
infrastructure robustness, flood and fire defenses, water management systems, and land-surface porosity.

\begin{figure}[ht]
\centering
\includegraphics[width=0.88\linewidth]{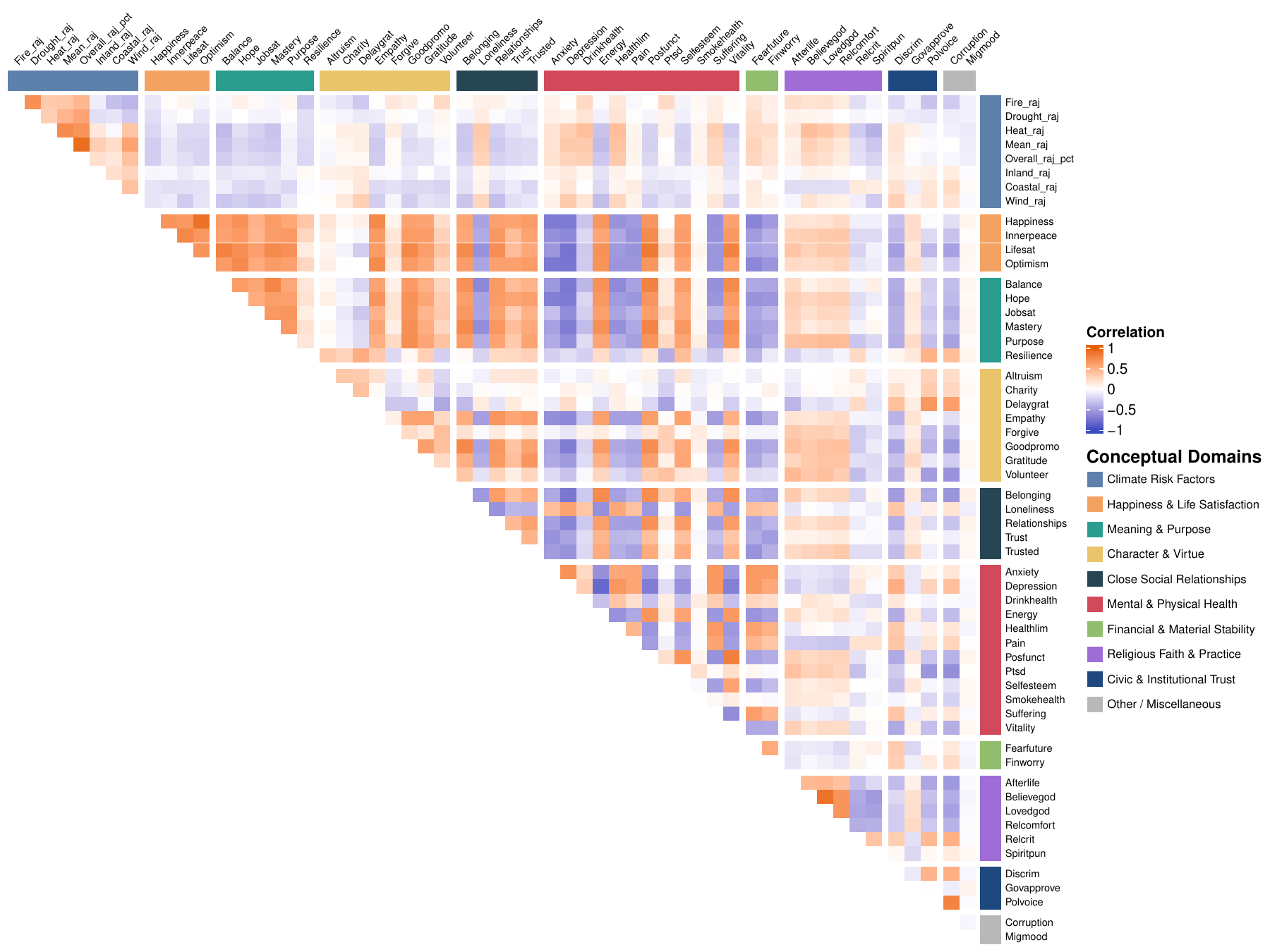}
\caption{Spearman rank correlations between county-level resilience-adjusted climate risk indicators and our social
human flourishing dimensions (average values by county 2013--2023).
Climate risk variables (\texttt{Fire\_raj}, \texttt{Drought\_raj}, \texttt{Heat\_raj},
\texttt{Inland\_raj}, \texttt{Coastal\_raj}, \texttt{Wind\_raj}, \texttt{Mean\_raj}, and
\texttt{Overall\_raj\_pct}) represent percentile-ranked resilience-adjusted hazard scores (0--100)
quantifying expected exposure to wildfire, drought, heat stress, inland and coastal flooding, and
hurricane wind after accounting for local adaptive capacity (see Table~\ref{tab:climate_risk_vars}).
The upper-triangular heatmap highlights distinct clusters: resilience-adjusted climate hazards form a
coherent block, while their correlations with flourishing domains are predominantly negative for
experiential and affective well-being (happiness, purpose, mental health) and near-zero or slightly
positive for moral, civic, and religious dimensions.}
\label{fig:corr_climate_flourishing}
\end{figure}

The correlation structure (Fig.~\ref{fig:corr_climate_flourishing}) reveals several consistent patterns
of association.  The resilience-adjusted climate variables form a cohesive cluster, indicating that
different hazard types tend to co-occur geographically and share common structural drivers of exposure.
Counties with higher composite or hazard-specific risk scores tend to exhibit lower levels of
\textit{Happiness and Life Satisfaction} (e.g., \textit{happiness}, \textit{innerpeace},
\textit{lifesat}, \textit{optimism}), \textit{Meaning and Purpose} (\textit{purpose},
\textit{jobsat}, \textit{resilience}, \textit{mastery}), and \textit{Mental and Physical Health}
(\textit{anxiety}, \textit{depression}, \textit{pain}, \textit{vitality}, \textit{energy},
\textit{healthlim}).  These negative rank correlations indicate that regions characterized by higher
resilience-adjusted risk (that is, greater expected hazard exposure relative to their adaptive capacity)
also tend to display weaker expressions of subjective well-being and vitality, although no causal
direction is implied.

By contrast, associations between resilience-adjusted climate risk and other flourishing domains appear
weaker or slightly positive.  Indicators of prosocial orientation and virtue (\textit{altruism},
\textit{charity}, \textit{volunteer}, \textit{gratitude}) and of religious sentiment
(\textit{believegod}, \textit{afterlife}, \textit{relcomfort}, \textit{lovedgod}) show near-zero or
mildly positive correlations with several risk measures.  This pattern indicates that moral and spiritual
discourse may remain stable, or occasionally more salient, in areas facing higher hazard exposure.
Similarly, civic and institutional trust indicators (\textit{govapprove}, \textit{polvoice}) display
small or heterogeneous associations, suggesting that political attitudes and confidence in governance
are less tightly coupled with environmental stress.

Overall, the observed associations show that resilience-adjusted climate risk correlates negatively with experiential and affective well-being, while its relationship with moral, social, and religious domains is weaker or neutral. 
These findings characterize patterns of co-variation between environmental hazard exposure and social expressions of flourishing. Any potential causal relationships remain beyond the scope of the present descriptive analysis.

\clearpage

\section{Usage Notes}

This section summarizes practical considerations for using the Human Flourishing Geographic Index (HFGI), including representativeness, spatial uncertainty, aggregation choices, measurement caveats, privacy safeguards, and appropriate use cases. All associations reported in this paper are descriptive; the dataset is not designed for causal inference.

\subsection{Representativeness and selection}
HFGI relies on geotagged tweets, which constitute roughly 1--2\% of all Twitter posts on average. As such, the dataset reflects the behavior of \emph{Twitter users who opt into geotagging}, not the general population. Prior work suggests geotagged posts can differ in affect from non-geotagged content and exhibit distinct usage contexts \cite{chai2023tsgi_scidata,Jiang2023_Carmen,Wang2022_COVIDSent}. Researchers should therefore treat HFGI as a measure of \emph{expressed} well-being among geotagging users rather than a direct proxy for population prevalence. Automated or organizational accounts were not explicitly filtered, because reliable, reproducible bot-detection methods are known to vary over time and across languages. Consequently, the HFGI reflects the aggregate expression visible in the public geotagged stream—human and institutional—rather than a demographically weighted population sample.

\paragraph{Practical guidance.}
When possible, (i) include controls for digital access and adoption (e.g., broadband coverage, device ownership), (ii) consider rural/urban status to account for connectivity and stigma-related expression differences, and (iii) report robustness to alternative normalizations. Demographic post-stratification is not feasible here (no reliable user age/sex), so HFGI indexes the expressed discourse of geotagging users.

\subsection{Spatial attribution and uncertainty}
Tweets are localized either by GPS or by a Twitter ``Place'' bounding box. For Place-only tweets, we use the centroid and record a spatial error derived from the box area. This introduces location uncertainty that is larger in sparsely populated regions and for large administrative places.

\paragraph{Practical guidance.}
Prefer county-level analyses (as provided) rather than finer granularities. For applications sensitive to precise localization, consider excluding records with large spatial error or conduct sensitivity analyses that weight observations inversely by spatial uncertainty.

\subsection{Platform policy changes over time}
Twitter location-sharing policies changed circa 2019, reducing the share of GPS-exact coordinates in favor of Place-based locations \cite{Zhang2022_Carmen2}. This can alter the composition and precision of geolocated content over time.

\paragraph{Practical guidance.}
Include year (or month) fixed effects in statistical models and examine pre/post-2019 sensitivity. 

\subsection{Aggregation choices, uncertainty, and salience}

Each flourishing dimension is classified per tweet with values in $\{-1,\,0,\,0.5,\,1\}$, where $-1$ indicates a negative expression of the dimension, $+0.5$ and $+1$ correspond to somewhat or clearly positive expressions, and $0$ denotes that the dimension is not expressed in the tweet.  
For each combination of \{geography, time period, dimension\}, the dataset reports the following key fields:

\begin{itemize}
  \item \texttt{stat}: the conditional mean of the coded values over tweets where the dimension is present (i.e., excluding zeros);
  \item \texttt{stat\_se}: the within-cell standard deviation of \texttt{stat}, reflecting the dispersion of the underlying tweet-level scores and thus the reliability of the cell estimate;
  \item \texttt{validtweets}: the number of tweets in which the given dimension was expressed;
  \item \texttt{ntweets}: the total number of tweets in the cell (all topics combined);
  \item \texttt{salience}: the share of applicable tweets, i.e.\ \texttt{validtweets/ntweets}, providing a measure of the relative prevalence of the dimension.
\end{itemize}

This structure allows users to assess both \emph{signal strength} (via \texttt{stat}) and \emph{estimate quality} (via \texttt{stat\_se}), while retaining tweet counts for replication weighting and uncertainty propagation.  
Because the conditional means are computed only over valid tweets, the resulting indicators are not biased toward zero by the inclusion of irrelevant content.

\paragraph{Practical guidance.}
\begin{enumerate}
  \item Treat \texttt{stat\_se} as a direct measure of uncertainty for each estimate. When aggregating or modeling indicators, use inverse-variance weighting:
  \[
  w_i = \frac{1}{\texttt{stat\_se}_i^{2}},
  \]
  or equivalently $w_i = 1 / (\texttt{stat}_i + \texttt{stat\_se}_i^{2})$ if combining information from both magnitude and precision.  
  This approach gives more influence to counties and periods where the estimate is statistically more stable.
  \item Apply minimum-volume filters, such as requiring $\texttt{validtweets} \ge x$ (where $x$ is a reasonable treshold for the geographical resolution), to ensure sufficient data support for each cell.
\end{enumerate}

\subsection{Expression vs.\ prevalence}
HFGI measures the \emph{propensity of public expression} for a concept, not clinical or survey prevalence. As shown in our external validations, social-media expression can diverge from survey prevalence in systematic ways (e.g., rural/urban gradients), consistent with communication and stigma theories. Apparent discrepancies can thus be informative about visibility, norms, and opportunity for expression.

\paragraph{Practical guidance.}
Interpret coefficients as \emph{associations with expressed content}. Where feasible, triangulate with independent sources (e.g., CDC PLACES for mental distress) and model both structural context (e.g., rurality) and their interactions.

\subsection{Privacy and ethical use}
No user-level identifiers are released; all outputs are aggregated at county or state level and by month or year, this guarantees enough privacy protection. Nevertheless, users of this dataset should avoid re-identification attempts and comply with platform terms and institutional review standards. The data are intended for research, not for profiling individuals or small groups.
All analyses are based on publicly available posts retrieved through the Twitter/X API.  Our work did not involve any interaction with individuals or collection of private data and therefore did not require human-subjects ethics review under institutional or U.S. federal guidelines.

\subsection{Recommended uses and cautions}
\textbf{Appropriate:} (i) descriptive mapping of spatial and temporal patterns; (ii) associational studies with external covariates; (iii) validation against independent indicators; (iv) policy monitoring where \emph{expression} is the target (e.g., public discourse around well-being).  
\textbf{Caution:} (i) causal inference from HFGI alone; (ii) small-area comparisons with low \texttt{validtweets}; (iii) cross-period comparisons without accounting for platform policy changes; (iv) interpreting expression metrics as clinical prevalence.

\subsection{Linking to exogenous exposures}
When linking HFGI to environmental or structural datasets (e.g., resilience-adjusted climate risk, economic indicators), ensure consistent spatial joins (county FIPS) and temporal alignment. All reported relationships are \emph{associational} unless using thoughtful causal models; HFGI should be interpreted as an outcome describing public expression conditioned by both underlying conditions and communication opportunity/propensity.

\medskip
In sum, HFGI offers high-resolution measures of expressed flourishing across space and time. With appropriate attention to representativeness, spatial uncertainty, aggregation, and the distinction between expression and prevalence, the dataset can support robust descriptive and associational analyses across disciplines.

\section*{Data Availability}
Data are available at Harvard Dataverse   at the link: \url{https://doi.org/10.7910/DVN/T39JBY} in both parquet and csv formats. More information can be found here \cite{HFGI}.

\section*{Data License}
This data is released under the Creative Commons CC0 1.0 Universal licence (\url{https://creativecommons.org/publicdomain/zero/1.0/}).

\section*{Code Availability}
Code for text processing, model fine-tuning, and statistics generation is available at: \url{https://github.com/siacus/flourishing-i-challenge} and related links.

\section*{Acknowledgments}
We acknowledge the support of Harvard FAS Research Computing for providing most of the computational resources, the NSF ACCESS program for providing initial computing bandwidth to start the analysis and Kempner Institute for the Study of Natural and Artificial Intelligence to allow us to use spare cycles of their GPU cluster. We thank CGA's Xiaokang Fu for his assistance in the data processing. We also thanks Parag Khanna for providing the {AlphaGeo~AI}  dataset on county-level resilience-adjusted climate risk indicators used in Section~\ref{sec:risk}.

\section*{Author contributions}
SI led the project and wrote the manuscript, with contributions from all authors. DJ provided code and data to extract the raw Geo Tweet Archive. SI and AN wrote the code to perform fine-tuning and running inference on GPU clusters. SI wrote the code to generate the HFGI indicators and the analyses in this manuscript. AN, GP, MC and AV  helped with theoretical framework and coordinated human coders.

\section*{Competing interests}
The authors declare no competing interests.

\bibliographystyle{unsrt}
\bibliography{hf_references}

\newpage
\eject

\section*{Supplementary Material}
\setcounter{table}{0}
\setcounter{figure}{0}
\renewcommand{\thetable}{S\arabic{table}}
\renewcommand{\thefigure}{S\arabic{figure}}

\begin{table}[htbp]
\centering
\caption{Data Schema of Geotweet Archive v2.0}
\label{tab:dbase}
\begin{tabular}{p{3.2cm} p{3cm} p{8.5cm}}
\hline
\textbf{Field Name} & \textbf{Type} & \textbf{Description} \\
\hline
\texttt{message\_id} & \texttt{BIGINT} & Tweet ID \\
\texttt{tweet\_date} & \texttt{TIMESTAMP} & Date and time of tweet from Twitter (UTC) \\
\texttt{tweet\_text} & \texttt{TEXT ENCODING} & Text content of tweet \\
\texttt{tags} & \texttt{TEXT ENCODING DICT} & Tweet hashtags \\
\texttt{tweet\_lang} & \texttt{TEXT ENCODING DICT} & Language of the tweet \\
\texttt{source} & \texttt{TEXT ENCODING DICT} & Operating system or application type used to create the tweet \\
\texttt{place*} & \texttt{TEXT ENCODING NONE} & User-defined geographic place, usually a town name. Twitter provides a bounding box, from which centroids (longitude/latitude) and spatial error are derived if GPS coordinates are not present. \\
\texttt{retweets} & \texttt{SMALLINT} & Number of retweets (as of last check) \\
\texttt{tweet\_favorites} & \texttt{SMALLINT} & Number of likes (formerly called ``favorites'') \\
\texttt{photo\_url} & \texttt{TEXT ENCODING DICT} & URL of any image referenced \\
\texttt{quoted\_status\_id} & \texttt{BIGINT} & ID number for quoted status \\
\texttt{user\_id} & \texttt{BIGINT} & User ID number \\
\texttt{user\_name} & \texttt{TEXT ENCODING NONE} & User name \\
\texttt{user\_location*} & \texttt{TEXT ENCODING NONE} & User-defined location, usually a city or town (see Twitter user object) \\
\texttt{followers} & \texttt{SMALLINT} & Number of followers (as of last check) \\
\texttt{friends} & \texttt{SMALLINT} & Number of users followed by this user \\
\texttt{user\_favorites} & \texttt{INT} & Number of topics the user is interested in \\
\texttt{status} & \texttt{INT} & Code for what the user is doing (as of last check) \\
\texttt{user\_lang} & \texttt{TEXT ENCODING DICT} & User-defined language \\
\texttt{latitude} & \texttt{FLOAT} & Latitude from GPS or bounding box (derived from Place field) \\
\texttt{longitude} & \texttt{FLOAT} & Longitude from GPS or bounding box (derived from Place field) \\
\texttt{data\_source*} & \texttt{TEXT ENCODING DICT} & Source crawler or dataset for the tweet \\
\texttt{gps} & \texttt{TEXT ENCODING DICT} & Flag for whether coordinates are from GPS or bounding box (SRID 4326). GPS takes priority when both are present. \\
\texttt{spatial\_error} & \texttt{FLOAT} & Horizontal error estimate (meters). 10m for GPS; for bounding box-derived coordinates, error is the radius of a circle with area equal to the bounding box. \\
\hline
\end{tabular}
\end{table}

\begin{table}[ht]
\centering
\caption{Tweet volume and U.S. share by year, 2010--2023.}
\label{tab:tweet_volume_us_share}
\begin{tabular}{lrrr}
\hline
\textbf{Year} & \textbf{U.S. Tweets} & \textbf{Total Tweets} & \textbf{U.S. Share (\%)} \\
\hline
2010 & 39 & 81 & 48.15 \\
2011 & 523 & 1,390 & 37.63 \\
2012 & 3{,}317{,}745 & 8{,}452{,}080 & 39.25 \\
2013 & 419{,}395{,}440 & 1{,}417{,}285{,}288 & 29.59 \\
2014 & 773{,}663{,}105 & 2{,}608{,}788{,}319 & 29.66 \\
2015 & 249{,}224{,}839 & 975{,}605{,}056 & 25.55 \\
2016 & 34{,}018{,}076 & 185{,}899{,}224 & 18.30 \\
2017 & 72{,}199{,}684 & 221{,}914{,}402 & 32.53 \\
2018 & 165{,}667{,}362 & 528{,}238{,}341 & 31.36 \\
2019 & 370{,}674{,}625 & 1{,}284{,}723{,}163 & 28.85 \\
2020 & 327{,}296{,}142 & 1{,}242{,}159{,}887 & 26.35 \\
2021 & 209{,}408{,}955 & 860{,}425{,}905 & 24.34 \\
2022 & 192{,}535{,}480 & 969{,}258{,}740 & 19.86 \\
2023 & 49{,}001{,}162 & 236{,}729{,}045 & 20.70 \\
\hline
\end{tabular}
\end{table}

\begin{table}[htbp]
\centering
\caption{Schema for \texttt{flourishingCountyMonth.(csv|parquet)} — County-level, monthly records (one row per \{county, year, month, variable\}).}
\label{tab:county_month_schema}
\begin{tabular}{p{3.3cm} p{2.6cm} p{9cm}}
\hline
\textbf{Field} & \textbf{Type} & \textbf{Description} \\
\hline
\texttt{variable} & \texttt{STRING} & Dimension name (same domain as County–Year). \\
\texttt{stat} & \texttt{FLOAT} & Monthly indicator in $[-1,+1]$ (conditional mean). \\
\texttt{stat\_se} & \texttt{FLOAT} & Standard error (if computed). \\
\texttt{salience} & \texttt{FLOAT} & Monthly share/proportion of applicable tweets (0–1). \\
\texttt{ntweets} & \texttt{INT} & Total tweets in the county–month cell. \\
\texttt{validtweets} & \texttt{INT} & Tweets where \texttt{variable} is present (denominator). \\
\texttt{natweets} & \texttt{INT} & Tweets where \texttt{variable} is absent. \\
\texttt{FIPS} & \texttt{STRING} & 2-digit state FIPS (zero-padded). \\
\texttt{county} & \texttt{STRING} & 3-digit county code (zero-padded). \\
\texttt{StateCounty} & \texttt{STRING} & 5-digit county FIPS (= \texttt{FIPS} $\Vert$ \texttt{county}). \\
\texttt{year} & \texttt{INT} & Calendar year. \\
\texttt{month} & \texttt{INT} & Calendar month (1–12). \\
\hline
\end{tabular}
\end{table}

\begin{table}[htbp]
\centering
\caption{Schema for \texttt{flourishingStateYear.(csv|parquet)} — State-level, yearly records (one row per \{state, year, variable\}).}
\label{tab:state_year_schema}
\begin{tabular}{p{3.3cm} p{2.6cm} p{9cm}}
\hline
\textbf{Field} & \textbf{Type} & \textbf{Description} \\
\hline
\texttt{variable} & \texttt{STRING} & Dimension name. \\
\texttt{stat} & \texttt{FLOAT} & Indicator value in $[-1,+1]$ (conditional mean). \\
\texttt{stat\_se} & \texttt{FLOAT} & Standard error (if computed). \\
\texttt{salience} & \texttt{FLOAT} & Share/proportion of applicable tweets (0–1). \\
\texttt{ntweets} & \texttt{INT} & Total tweets in the state–year cell. \\
\texttt{validtweets} & \texttt{INT} & Tweets where \texttt{variable} is present. \\
\texttt{natweets} & \texttt{INT} & Tweets where \texttt{variable} is absent. \\
\texttt{FIPS} & \texttt{STRING} & 2-digit state FIPS (zero-padded). \\
\texttt{state} & \texttt{STRING} & State name or postal abbreviation (if present). \\
\texttt{year} & \texttt{INT} & Calendar year. \\
\hline
\end{tabular}
\end{table}

\begin{table}[htbp]
\centering
\caption{Schema for \texttt{flourishingStateMonth.(csv|parquet)} — State-level, monthly records (one row per \{state, year, month, variable\}).}
\label{tab:state_month_schema}
\begin{tabular}{p{3.3cm} p{2.6cm} p{9cm}}
\hline
\textbf{Field} & \textbf{Type} & \textbf{Description} \\
\hline
\texttt{variable} & \texttt{STRING} & Dimension name. \\
\texttt{stat} & \texttt{FLOAT} & Monthly indicator in $[-1,+1]$ (conditional mean). \\
\texttt{stat\_se} & \texttt{FLOAT} & Standard error (if computed). \\
\texttt{salience} & \texttt{FLOAT} & Monthly share/proportion of applicable tweets (0–1). \\
\texttt{ntweets} & \texttt{INT} & Total tweets in the state–month cell. \\
\texttt{validtweets} & \texttt{INT} & Tweets where \texttt{variable} is present. \\
\texttt{natweets} & \texttt{INT} & Tweets where \texttt{variable} is absent. \\
\texttt{FIPS} & \texttt{STRING} & 2-digit state FIPS (zero-padded). \\
\texttt{state} & \texttt{STRING} & State name or postal abbreviation (if present). \\
\texttt{year} & \texttt{INT} & Calendar year. \\
\texttt{month} & \texttt{INT} & Calendar month (1–12). \\
\hline
\end{tabular}
\end{table}

\begin{figure}[ht]
\centering
\includegraphics[width=0.88\linewidth]{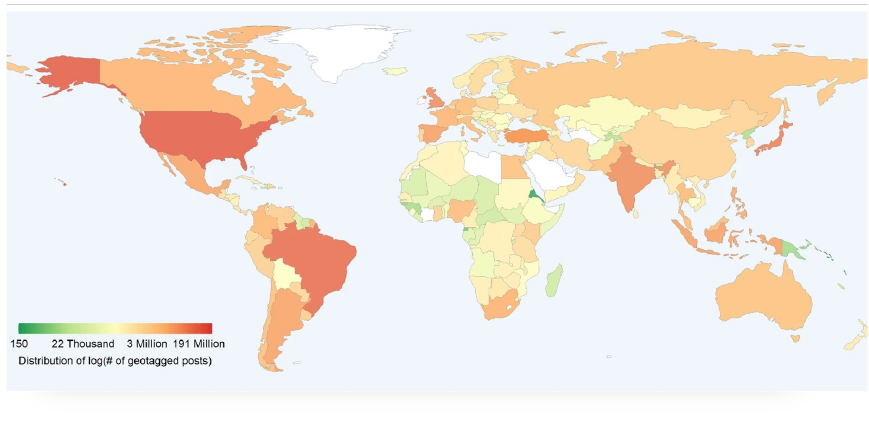}
\caption{The world map illustrates the spatial distribution of collected geotagged posts at the country level for 2021 \cite{chai2023tsgi_scidata, hfgs_dataset}.}
\label{fig:worldmap}
\end{figure}

\begin{figure}[ht]
\centering
\includegraphics[width=0.88\linewidth]{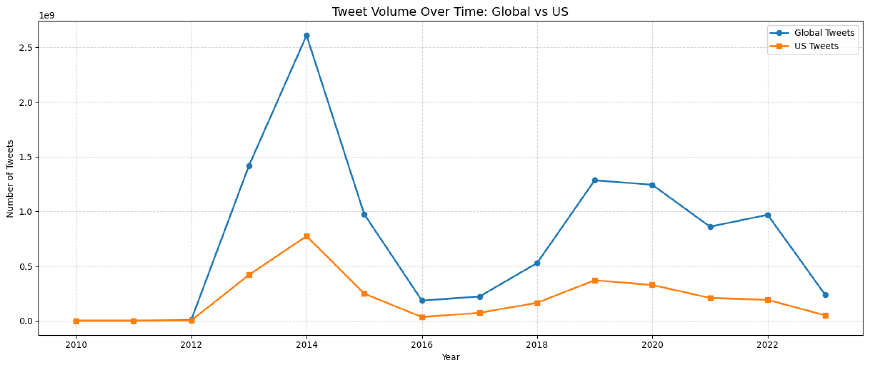}
\caption{Tweet volume over time: Global vs.\ United States.}
\label{fig:timeseries}
\end{figure}

\begin{table}[htbp]
\centering
\caption{List of the 46 flourishing dimensions used in the Human Flourishing Geographic Index dataset. Variables correspond to the column \texttt{stat} in the dataset tables (County/State, Monthly/Yearly) and their GFS related topic. In addition to that we have corruption and migration mood for a total of 48 dimensions.}
\label{tab:flourishing_dimensions}
\small
\begin{tabular}{p{4cm} p{9.5cm}}
\hline
\textbf{Variable name} & \textbf{Description (plain English)} \\
\hline
\texttt{afterlife} & Life after death belief \\
\texttt{altruism} & Expressing altruism \\
\texttt{anxiety} & Anxiety \\
\texttt{balance} & Balance in the various aspects of own life \\
\texttt{believegod} & Belief in God \\
\texttt{belonging} & Belonging to society \\
\texttt{charity} & Charitable giving or helping \\
\texttt{delaygrat} & Expressing delayed gratification \\
\texttt{depression} & Depression \\
\texttt{discrim} & Perceiving discrimination \\
\texttt{drinkhealth} & Describing drinking related health issues \\
\texttt{empathy} & Expressing empathy \\
\texttt{energy} & Having energy \\
\texttt{fearfuture} & Fear of future \\
\texttt{finworry} & Financial or material worry \\
\texttt{forgive} & Seeking for forgiveness \\
\texttt{goodpromo} & Promoting good \\
\texttt{govapprove} & Expressing government approval \\
\texttt{gratitude} & Expressing gratitude \\
\texttt{happiness} & Happiness \\
\texttt{healthlim} & Describing health limitations \\
\texttt{hope} & Having hope \\
\texttt{innerpeace} & Peace with thoughts and feelings \\
\texttt{jobsat} & Expressing job satisfaction \\
\texttt{lifesat} & Life satisfaction \\
\texttt{lovedgod} & Feeling loved by God \\
\texttt{loneliness} & Loneliness \\
\texttt{mastery} & Mastery (ability or capability) \\
\texttt{optimism} & Expressing optimism \\
\texttt{pain} & Feeling pain \\
\texttt{polvoice} & Feeling having a political voice \\
\texttt{posfunct} & Positive functioning \\
\texttt{ptsd} & PTSD (post-traumatic stress disorder) \\
\texttt{purpose} & Purpose in life \\
\texttt{relationships} & Quality of relationships \\
\texttt{relcomfort} & Feeling religious comfort \\
\texttt{relcrit} & Religious criticism \\
\texttt{resilience} & Resilience \\
\texttt{selfesteem} & Self-esteem \\
\texttt{smokehealth} & Describing smoking related health issues \\
\texttt{spiritpun} & Spiritual punishment \\
\texttt{suffering} & Suffering \\
\texttt{trust} & Expressing trust \\
\texttt{trusted} & Feeling trusted \\
\texttt{vitality} & Vitality \\
\texttt{volunteer} & Volunteering \\
\hline
\texttt{corruption} & Perception of corruption \\
\texttt{migmood} & Mood towards migration/migrants \\
\end{tabular}
\end{table}

\begin{table}[htbp]
\centering
\caption{Approximate mapping of some of the  flourishing dimensions used in the Human Flourishing Geographic Index dataset and the original questions in the GFS survey.}
\label{tab:flourishing_GFS}
\small
\begin{tabular}{p{2.5cm} p{12.5cm}}
\hline
\textbf{HFGI Var.} & \textbf{GFS survey question} \\
\hline
\texttt{afterlife} & Do you believe in life after death, or not?  \\
\texttt{anxiety} & Over the last 2 weeks, how often have you been bothered by the following problems? 
Feeling nervous, anxious or on edge;
Not being able to stop or control worrying
 \\
\texttt{balance} &  In general, how often are the various aspects of your life in balance? \\
\texttt{believegod} &  Do you believe in one God, more than one god, an impersonal spiritual force, or none of these? 
 \\
\texttt{belonging} &   How would you describe your sense of belonging
to your local community?\\
\texttt{charity} &  In the past month, have you donated money to a charity?  \\
\texttt{delaygrat} & I am always able to give up some happiness now for greater happiness later. 
  \\
\texttt{depression} & Over the last 2 weeks, how often have you been bothered by the following problems? Little interest or pleasure in doing things; Feeling down, depressed or hopeless
 \\
\texttt{discrim} &  How often do you feel discriminated against because of any group you are a part of? This might include discrimination because of your religion, political affiliation, race, gender, social class, sexual orientation, or involvement in civic organizations or community groups. 
 \\
\texttt{fearfuture} &  Just your best guess, on which step do you think you will stand in the future, say about five years from now? [Worst possible - Best possible] \\
\texttt{finworry} &  How often do you worry about being able to meet normal monthly living expenses? 
How often do you worry about safety, food, or housing? 
 \\
\texttt{forgive} & How often have you forgiven those who have hurt you?  \\
\texttt{goodpromo} & I always act to promote good in all circumstances, even in difficult and challenging situations. 
  \\
\texttt{govapprove} &  How much do you approve or disapprove of the job performance of the national government of this country? 
 \\
\texttt{gratitude} & If I had to list everything that I felt grateful for, it would be a very long list.  \\
\texttt{happiness} &  In general, how happy or unhappy do you usually feel? \\
\texttt{healthlim} & Do you have any health problems that prevent you from doing any of the things people your age normally can do?  \\
\texttt{hope} &  Despite challenges, I always remain hopeful about the future. \\
\texttt{innerpeace} & In general, how often do you feel you are at peace with your thoughts and feelings? 
\\
\texttt{lifesat} & Overall, how satisfied are you with life as a whole these days?  \\
\texttt{lovedgod} & I feel loved or cared for by God, the main god I worship, or the spiritual force that guides my life.\\
\texttt{loneliness} & How often do you feel lonely? 
 \\
\texttt{mastery} & How often do you feel very capable in most things you do in life? \\
\texttt{optimism} & Overall, I expect more good things to happen to me than bad. 
  \\
\texttt{pain} &  How much bodily pain have you had during the past 4 weeks?\\
\texttt{polvoice} & Do you agree or disagree with the following statement? People like me have a say about what the government does. 
\\
\texttt{ptsd} &  Think about the biggest threat to life you've ever witnessed or experienced first-hand during your life. In the past month, how much have you been bothered by this experience? \\
\texttt{purpose} &  Overall, to what extent do you feel the things you do in your life are worthwhile? 
\\
\texttt{relationships} & My relationships are as satisfying as I would want them to be. 
 \\
\texttt{relcomfort} & I find strength or comfort in my religion
or spirituality. \\
\texttt{relcrit} & People in my religious community are critical of me or my lifestyle. 
 \\
\texttt{volunteer} &  In the past month, have you volunteered your time to an organization?\\
\end{tabular}
\end{table}

\begin{table*}[ht]
\centering
\caption{\textbf{Climate risk variables and descriptions.}
All risk variables are percentile scores (0–100) derived from the
\textit{Climate Risk and Resilience-adjusted Index} (2005--2100), where higher values
indicate greater exposure to physical climate hazards after accounting for local resilience factors.}
\label{tab:climate_risk_vars}
\begin{tabular}{p{3cm} p{5cm} p{7cm}}
\hline
\textbf{Variable} & \textbf{Expanded Name} & \textbf{Description} \\
\hline
\texttt{FIRE\_RAJ} &
Resilience-adjusted Wildfire Risk Score &
Composite percentile (0--100) measuring exposure to wildfire hazard, incorporating
the effectiveness of local fire monitoring systems and defense infrastructure.
A higher score indicates greater wildfire risk. \\[3pt]

\texttt{DROUGHT\_RAJ} &
Resilience-adjusted Drought Risk Score &
Percentile-based drought exposure index integrating the capacity of local water supply
systems and emergency water infrastructure.
Higher values correspond to greater resilience-adjusted drought risk. \\[3pt]

\texttt{HEAT\_RAJ} &
Resilience-adjusted Heat Stress Risk Score &
Percentile rank reflecting resilience-adjusted exposure to heat stress,
incorporating Urban Heat Island effects such as building density and green cover. \\[3pt]

\texttt{INLAND\_RAJ} &
Resilience-adjusted Inland Flooding Risk Score &
Composite percentile of inland flooding risk that includes surface porosity, flood
control measures, and defensive infrastructure against riverine inundation. \\[3pt]

\texttt{COASTAL\_RAJ} &
Resilience-adjusted Coastal Flooding Risk Score &
Percentile-based index reflecting exposure to coastal flooding and sea-level rise,
adjusted for the presence of coastal flood defenses and protection infrastructure. \\[3pt]

\texttt{WIND\_RAJ} &
Resilience-adjusted Hurricane Wind Risk Score &
Percentile risk score representing hurricane-related wind exposure, incorporating
building strength and local storm mitigation capacity. \\[3pt]

\texttt{MEAN\_RAJ} &
Mean Resilience-adjusted Climate Risk Score &
Average of all hazard-specific resilience-adjusted risk scores (heat, drought, inland
and coastal flooding, wind, and wildfire). Represents overall multi-hazard exposure. \\[3pt]

\texttt{OVERALL\_RAJ\_PCT} &
Overall Climate Resilience-adjusted Risk Percentile &
Global percentile rank summarizing total climate hazard exposure after adjusting
for adaptation capacity. Higher values indicate greater overall climate vulnerability. \\[3pt]
\hline
\end{tabular}
\end{table*}

\clearpage

\eject

\section{Codebook: Well-Being Dimensions (Twitter-based Classification)}
Within this research project, we approach the concept of \emph{human flourishing} as  a multifaceted concept. For more information see \cite{vanderweele2017promotion, gfs}.

\subsection{Tracing population well-being using Twitter/X}
We operationalize the distinctive features of well-being and trace them in people’s tweets through a large-language-model (LLM) classifier. For each tweet, applicable dimensions (from the list below) are identified and assigned an intensity label \texttt{low}, \texttt{medium}, or \texttt{high}. Non-applicable dimensions are omitted. In summary:
\begin{itemize}[nosep]
  \item use \textbf{low}: negative or opposite expression of the construct;
  \item use \textbf{medium}: moderate/ambiguous/partial alignment to the construct;
  \item use \textbf{high}: clear and strong alignment to the construct;
  \item a tweet may map to \textbf{multiple} dimensions;
  \item non-applicable dimensions are omitted and should  never be coded as ``low''.
\end{itemize}

\subsection{Identifying the Human Flourishing Dimensions}
This section defines each dimension and provides example tweets with the intended label. All examples are synthetic or heavily paraphrased to protect Twitter accounts' privacy.

\setlist[description]{ leftmargin=0pt, labelsep=0.5em}
\begin{description}

\item[1) Happiness.] The text expresses some level of happiness (\texttt{happiness}).\\
\textbf{low:} ``Mondays always feel endless and gray.'' \\
\textbf{med:} ``Trying to notice at least one small thing that makes me smile today.'' \\
\textbf{high:} ``We reached our team goal today—feeling genuinely proud and joyful!''

\item[2) Resilience.] Capability of withstanding or recovering from difficulties (\texttt{resilience}).\\
\textbf{low:} ``Nothing seems to help me move on from this.'' \\
\textbf{med:} ``It’s been a hard season, but we’re still finding ways to keep going.'' \\
\textbf{high:} ``Every setback is just another chance to start stronger than before.'' 

\item[3) Self-esteem.] Confidence in one's worth or abilities. (\texttt{selfesteem})\\
\textbf{low:} ``I never feel good enough for anyone.'' \\
\textbf{med:} ``I might not tick every box, but I know I bring value.'' \\
\textbf{high:} ``Feeling confident and loving who I’m becoming.'' 

\item[4) Life Satisfaction.] Satisfaction with one's life as a whole (\texttt{lifesat}).\\
\textbf{low:} ``My living situation is exhausting and unfair.'' \\
\textbf{med:} ``Things are slowly falling into place—can’t complain too much.'' \\
\textbf{high:} ``Watching the sunset from my porch tonight, feeling truly content.'' 

\item[5) Fear of future.] Worry about one's condition in the coming years (\texttt{fearfuture}).\\
\textbf{low:} ``I’m sure everything will work out somehow.'' \\
\textbf{med:} ``Uncertain about how next year will turn out financially.'' \\
\textbf{high:} ``The future feels like a cliff I’m walking toward in the dark.'' 

\item[6) Vitality.] Feelings of strength and activity (\texttt{vitality}).\\
\textbf{low:} ``Still recovering from a cold, running on empty.'' \\
\textbf{med:} ``Tired but determined to finish this week strong.'' \\
\textbf{high:} ``The energy today is unreal—everything feels possible!''

\item[7) Having energy.] Reports of feeling (or not) energetic (\texttt{energy}).\\
\textbf{low:} ``Can’t seem to get moving this morning.'' \\
\textbf{med:} ``A little sluggish but getting things done.'' \\
\textbf{high:} ``So full of energy I could run another mile!''

\item[8) Positive functioning.] Feeling capable and effective (\texttt{posfunct}).\\
\textbf{low:} ``Everything feels like too much lately.'' \\
\textbf{med:} ``Getting back into a productive rhythm again.'' \\
\textbf{high:} ``Handled a dozen tasks today—completely in the zone!''

\item[9) Expressing job satisfaction.] Satisfaction with one's present job (\texttt{jobsat}).\\
\textbf{low:} ``Work feels meaningless lately.'' \\
\textbf{med:} ``Busy week but I’m learning new things.'' \\
\textbf{high:} ``I truly enjoy what I do—it fits me perfectly.'' 

\item[10) Expressing optimism.] Optimism about one’s condition in the medium run (\texttt{optimism}).\\
\textbf{low:} ``I don’t expect much to improve soon.'' \\
\textbf{med:} ``Maybe next month will bring better news.'' \\
\textbf{high:} ``Things are finally turning around—I can feel it!'' 

\item[11) Peace with thoughts and feelings.] General feeling of calm acceptance (\texttt{innerpeace}).\\
\textbf{low:} ``Can’t quiet my thoughts enough to rest.'' \\
\textbf{med:} ``Learning to take things as they come.'' \\
\textbf{high:} ``Feeling calm and centered after a long day.'' 

\item[12) Purpose in life.] Acting with a sense of purpose (\texttt{purpose}).\\
\textbf{low:} ``I’m not sure what I’m doing any of this for.'' \\
\textbf{med:} ``Trying to find direction and meaning in small steps.'' \\
\textbf{high:} ``Every action today felt aligned with what I believe in.'' 

\item[13) Depression.] Expressions of hopelessness or anhedonia (\texttt{depression}).\\
\textbf{low:} ``Nothing negative—just a regular day.'' \\
\textbf{high:} ``Everything feels heavy and empty at once.'' 

\item[14) Anxiety.] Feeling nervous or unable to control worry (\texttt{anxiety}).\\
\textbf{low:} ``Calm and relaxed morning.'' \\
\textbf{high:} ``My heart won’t stop racing over things I can’t fix.'' 

\item[15) Suffering.] Expression of physical or mental pain (\texttt{suffering}).\\
\textbf{high:} ``The pain won’t let me sleep tonight.'' 

\item[16) Feeling pain.] Explicit mention of pain or grief (\texttt{pain}).\\
\textbf{high:} ``Missing someone so deeply it physically hurts.'' 

\item[17) Expressing altruism.] Acting for others’ benefit (\texttt{altruism}).\\
\textbf{low:} ``Wishing people well but staying uninvolved.'' \\
\textbf{med:} ``I hope things get better for you soon.'' \\
\textbf{high:} ``Spent the evening delivering meals to neighbors in need.'' 

\item[18) Loneliness.] Expressions of loneliness (\texttt{loneliness}).\\
\textbf{low:} ``Remembering old friends with fondness.'' \\
\textbf{med:} ``Sometimes being alone feels peaceful, sometimes not.'' \\
\textbf{high:} ``It’s like the whole world moved on without me.'' 

\item[19) Quality of relationships.] Satisfaction with relationships (\texttt{relationships}).\\
\textbf{low:} ``We argue more than we talk lately.'' \\
\textbf{med:} ``Dinner with family went better than expected.'' \\
\textbf{high:} ``I’m grateful every day for the people who truly care.'' 

\item[20) Belonging to society.] Sense of community belonging (\texttt{belonging}).\\
\textbf{low:} ``Feels like I don’t fit anywhere anymore.'' \\
\textbf{high:} ``Proud of my community for showing up today!'' 

\item[21) Expressing gratitude.] Feeling and expressing gratitude (\texttt{gratitude}).\\
\textbf{high:} ``So thankful for the help and kindness I received today.'' 

\item[22) Expressing trust.] Trust toward people or institutions (\texttt{trust}).\\
\textbf{low:} ``Hard to rely on anyone lately.'' \\
\textbf{med:} ``I’ll give them another chance.'' \\
\textbf{high:} ``I know they’ll do the right thing.'' 

\item[23) Feeling trusted.] Reporting that one is trusted by others (\texttt{trusted}).\\
\textbf{low:} ``They still don’t believe in me.'' \\
\textbf{high:} ``Being asked for advice feels good—it means they trust me.'' 

\item[24) Balance in life.] Maintaining equilibrium across life domains (\texttt{balance}).\\
\textbf{low:} ``Work completely took over my week.'' \\
\textbf{med:} ``Still learning to leave the laptop closed after 6.'' \\
\textbf{high:} ``Spent the morning running and the evening with friends—perfect balance.'' 

\item[25) Mastery (ability/capability).] Feeling competent (\texttt{mastery}).\\
\textbf{low:} ``Everything I try seems to go wrong.'' \\
\textbf{med:} ``Finally getting the hang of this project.'' \\
\textbf{high:} ``Solved a tricky problem today—felt amazing!'' 

\item[26) Perceiving discrimination.] Feeling discriminated against (\texttt{discrim}).\\
\textbf{med:} ``People keep mispronouncing my name—it gets tiring.'' \\
\textbf{high:} ``I was turned away just because of where I’m from.'' 

\item[27) Feeling loved by God.] Feeling loved by a higher power (\texttt{lovedgod}).\\
\textbf{high:} ``I feel surrounded by divine care and peace.'' 

\item[28) Belief in God.] Belief in a divine entity (\texttt{believegod}).\\
\textbf{med:} ``Faith gives me comfort even when I doubt.'' \\
\textbf{high:} ``Trusting completely in God’s plan today.'' 

\item[29) Religious criticism.] Feeling criticized for one’s beliefs (\texttt{relcrit}).\\
\textbf{high:} ``They mocked my faith again, but I won’t hide it.'' 

\item[30) Spiritual punishment.] Feeling punished by a spiritual force (\texttt{spiritpun}).\\
\textbf{high:} ``Maybe this hardship is a test from above.'' 

\item[31) Feeling religious comfort.] Finding comfort in spirituality.\\
\textbf{high:} ``Prayer always brings me back to calm and strength.'' 

\item[32) Financial/material worry.] Concern about finances (\texttt{finworry}).\\
\textbf{low:} ``Joking about how expensive coffee is today.'' \\
\textbf{high:} ``Not sure how I’ll cover rent this month.'' 

\item[33) Life after death belief.] Belief in life after death (\texttt{afterlife}).\\
\textbf{high:} ``I believe our stories continue beyond this world.'' 

\item[34) Volunteering.] Giving time for others (\texttt{volunteer}).\\
\textbf{high:} ``Spent the weekend mentoring kids at the community center.'' 

\item[35) Charitable giving/helping.] Donating or assisting others (\texttt{charity}).\\
\textbf{high:} ``Joined a fundraiser—every bit helps someone in need.'' 

\item[36) Seeking forgiveness.] Willingness to forgive (\texttt{forgive}).\\
\textbf{high:} ``I’ve decided to let go and forgive—it feels lighter.'' 

\item[37) Feeling having a political voice.] Feeling empowered to influence politics (\texttt{polvoice}).\\
\textbf{med:} ``Signed a petition today—not sure it matters, but I tried.'' \\
\textbf{high:} ``Spoke at a town meeting—change starts small but real.'' 

\item[38) Expressing government approval.] Expressing satisfaction with governance (\texttt{govapprove}).\\
\textbf{high:} ``The new policy finally addresses what we’ve been asking for.'' 

\item[39) Having hope.] Hope about the future (\texttt{hope}).\\
\textbf{low:} ``Hard to see a way forward.'' \\
\textbf{med:} ``Things might improve if we stay patient.'' \\
\textbf{high:} ``Tomorrow will be brighter—I truly believe it.'' 

\item[40) Promoting good.] Acting to promote the common good (\texttt{goodpromo}).\\
\textbf{high:} ``Planted trees with neighbors today—small steps for a greener town.'' 

\item[41) Expressing delayed gratification.] Willingness to wait for future reward (\texttt{delaygrat}).\\
\textbf{low:} ``Can’t resist buying it now.'' \\
\textbf{med:} ``Saving a bit each week for the trip.'' \\
\textbf{high:} ``Working hard today for the long-term goal ahead.'' 

\item[42) PTSD.] Bothered by past traumatic experiences (\texttt{ptsd}).\\
\textbf{med:} ``Still have flashbacks sometimes.'' \\
\textbf{high:} ``Therapy is helping me process what happened years ago.'' 

\item[43) Smoking-related health issues.] Mentions of smoking and health (\texttt{smokehealth}).\\
\textbf{high:} ``Trying again to quit—breathing feels tougher lately.'' 

\item[44) Drinking-related health issues.] Mentions of alcohol use or effects (\texttt{drinkhealth}).\\
\textbf{med:} ``Had a few drinks too many, need to slow down.'' \\
\textbf{high:} ``Realizing drinking is hurting my health—it’s time to change.'' 

\item[45) Health limitations.] Health problems restricting daily activity (\texttt{healthlim}).\\
\textbf{low:} ``Catching a small cold, nothing serious.'' \\
\textbf{med:} ``Back pain keeps me from sitting too long.'' \\
\textbf{high:} ``Severe migraine kept me in bed all day.'' 

\item[46) Expressing empathy.] Showing compassion for others (\texttt{empathy}).\\
\textbf{low:} ``Everyone deals with their own stuff—don’t ask me to care.'' \\
\textbf{med:} ``That story was moving; hope they recover soon.'' \\
\textbf{high:} ``My heart breaks for the families affected—I wish I could help.'' 
\end{description}

\subsubsection{Additional coding rules for the sentiment towards migration and perception of corruption}

\setlist[description]{ leftmargin=0pt, labelsep=0.5em}
\begin{description}

\item[47) Sentiment towards Migration.] 
The text refers to attitudes or opinions about migration (\texttt{migmood}). Use these labels: ``pro-immigration'', ``anti-immigration'', ``neutral'', or ``unrelated''.

\textbf{pro-immigration:} ``Our community is stronger because people from many countries have made it their home.''\\
\textbf{anti-immigration:} ``Too many newcomers are taking jobs that should go to locals first.''\\
\textbf{neutral:} ``The city council is debating a new migration policy next week.''

\item[48) Corruption.] 
Determine whether the tweet is about \texttt{corruption}. Possible answers: ``yes'', ``no''.

\textbf{yes:} ``Another mayor accused of taking bribes, when will this end?''\\
\textbf{no:} ``Traffic downtown is unbearable today; everyone’s late to work.''

\end{description}

\end{document}